\documentclass[journal]{IEEEtran}

\makeatletter
\newcommand\notsotiny{\@setfontsize\notsotiny\@vipt\@viipt}
\makeatother

\usepackage{cite}
\usepackage{amsmath,amssymb,amsfonts}
\usepackage{stackrel}
\usepackage{algorithmic}
\usepackage{adjustbox}
\usepackage{graphicx}
\usepackage{textcomp}
\usepackage{xcolor}
\usepackage{float}
\usepackage[normalem]{ulem}
\usepackage{caption}    %
\usepackage{subcaption}    %
\usepackage{interval}
\usepackage{multicol, multirow}
\usepackage{siunitx}
\usepackage{booktabs}
\usepackage{pgfplots}
\usepackage[utf8]{inputenc}
\DeclareUnicodeCharacter{2212}{−}
\usepgfplotslibrary{groupplots,dateplot,colorbrewer, fillbetween}
\usetikzlibrary{patterns,shapes.arrows}
\pgfplotsset{compat=newest}

\def\BibTeX{{\rm B\kern-.05em{\sc i\kern-.025em b}\kern-.08em
    T\kern-.1667em\lower.7ex\hbox{E}\kern-.125emX}}

\def\synth{\emph{Sim}}

\newif\ifclearsectionlook

\newif\ifarxiv
\arxivtrue %

\global\long\def\MTF{\text{MTF}}
\global\long\def\PSF{\text{PSF}}
\global\long\def\PhTF{\text{PhTF}}
\global\long\def\smallSpace{\,}
\global\long\def\OTF{\text{OTF}}
\global\long\def\focal{\mathrm{f}}

\DeclareMathOperator*{\argmax}{arg\,max}

\DeclareSIUnit\px{px}
\DeclareSIUnit\lines{lines}
\DeclareSIUnit\patch{patch}
\DeclareSIUnit\DN{DN}

\definecolor{visioGreen}{RGB}{112, 173, 71}
\definecolor{visioBlue}{RGB}{70, 114, 196}
\definecolor{visioRed}{RGB}{255, 0, 0}

\usepackage{balance}
\usepackage[hyphens]{url}

\def\MyTitle{Monitoring and Adapting the Physical State of a Camera for Autonomous Vehicles} %

\newcommand\MYhyperrefoptions{bookmarks=true,bookmarksnumbered=true,
pdfpagemode={UseOutlines},plainpages=false,pdfpagelabels=true,
colorlinks=true,citecolor={black},
pdftitle={\MyTitle},%
pdfsubject={Image Processing, Automation, Robotics, Computer Vision},%
pdfauthor={M. Wischow, G. Gallego, I. Ernst, A. Börner},%
pdfkeywords={Condition Monitoring, Sensor AI, Active Vision, Image Blur, Image Noise, Camera Condition}}%
\usepackage[\MYhyperrefoptions,pdftex]{hyperref}

\ifarxiv
\usepackage[absolute]{textpos}
\fi

\begin{document}
\title{\MyTitle} 

\ifarxiv
\definecolor{somegray}{gray}{0.6}
\newcommand{\darkgrayed}[1]{\textcolor{somegray}{#1}}
\begin{textblock}{11}(2.5, 0.4)
\begin{center}
\darkgrayed{This paper has been accepted for publication at the\\
IEEE Transactions on Intelligent Transportation Systems, 2023.
\copyright IEEE}
\end{center}
\end{textblock}
\fi 

\author{Maik Wischow$^{1,2}$, Guillermo Gallego$^{2}$, %
Ines Ernst$^{1}$, Anko Börner$^{1}$%
\thanks{$^{1}$M.W., I.E. and A.B. are with the German Aerospace Center (DLR), Berlin, Germany. E-Mail: [firstname].[lastname]@dlr.de.}%
\thanks{$^{2}$G.G. is with 
the Dept. of EECS of TU Berlin (Faculty IV),
the Einstein Center Digital Future, and 
the Science of Intelligence Excellence Cluster, Berlin, Germany. 
E-Mail: guillermo.gallego@tu-berlin.de.}%
\thanks{Preprint of IEEE T-ITS paper. DOI: 10.1109/TITS.2023.3328811}%
}

\maketitle

\begin{abstract}
Autonomous vehicles and robots require increasingly more robustness and reliability to meet the demands of modern tasks.
These requirements specially apply to cameras onboard such vehicles because they are the predominant sensors to acquire information about the environment and support actions.
Cameras must maintain proper functionality and take automatic countermeasures if necessary. 
Existing solutions are typically tailored to specific problems or detached from the downstream computer vision tasks of the machines, which, however, determine the requirements on the quality of the produced camera images.
We propose a generic and task-oriented self-health-maintenance framework for cameras based on data- and physically-grounded models. 
To this end, we determine two reliable, real-time capable estimators for typical image effects of a camera in poor condition (blur, noise phenomena and most common combinations) by evaluating traditional and customized machine learning-based approaches in extensive experiments. 
Furthermore, we implement the framework on a real-world ground vehicle and demonstrate how a camera can adjust its parameters to counter an identified poor condition to achieve optimal application capability based on experimental (non-linear and non-monotonic) input-output performance curves. 
Object detection is chosen as target application, and the image effects motion blur and sensor noise as conditioning examples.
Our framework not only provides a practical ready-to-use solution to monitor and maintain the health of cameras, but can also serve as a basis for extensions to tackle more sophisticated problems that combine additional data sources (e.g., sensor or environment parameters) empirically in order to attain fully reliable and robust machines.
\end{abstract}


\IEEEpeerreviewmaketitle
\ifclearsectionlook\cleardoublepage\fi  \begin{figure*}[ht]
    \centering
    \includegraphics[width=.98\linewidth, trim=0 0 0 0, clip]{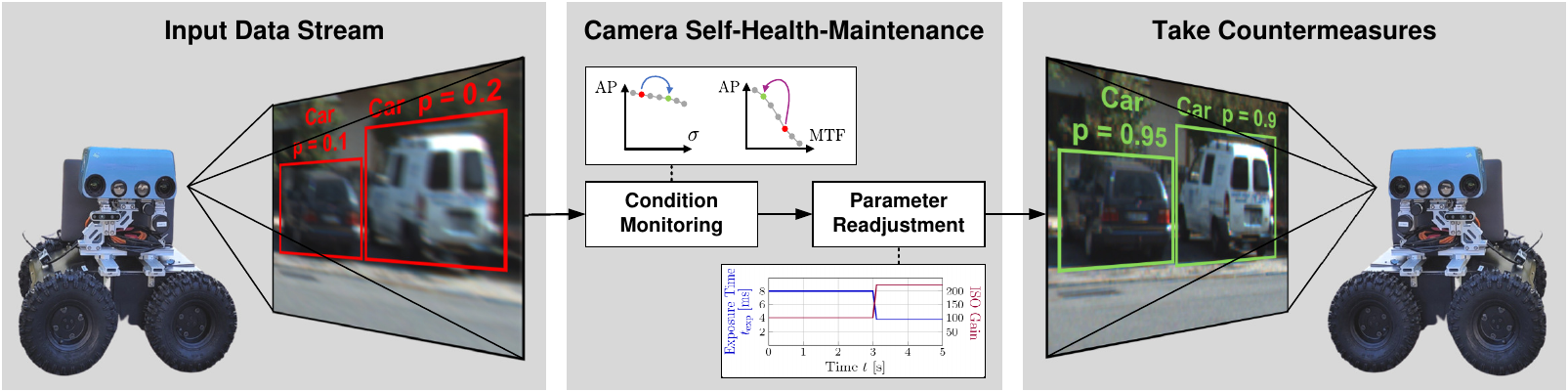}
    \caption{The ground vehicle fails to detect the motion blurred cars ({\color{visioRed}red}) given its current camera configuration. 
    We tackle the source of this problem using %
    ($i$) an online estimation of image quality properties,
    ($ii$) knowledge about camera physics and
    ($iii$) empirical object detection performance curves $\mathrm{AP}$ (expressed as functions of the image quality).
    In this way, unfavorable camera conditions can be detected and actively tackled to reach optimal application performance ({\color{visioGreen} green}). 
    In the example, image blur is estimated and mitigated online by changing the camera configuration: 
    decreasing the exposure time {\color{blue!80!black}$t_\text{exp}$} and increasing the {\color{violet!80!black}ISO} gain.
    Blur is reduced at the expense of slightly increasing noise to produce better object detection rates.
    }
    \label{fig:eyeCatcher}
\end{figure*}

\section{Introduction}
\label{sec:introduction}

\IEEEPARstart{M}{achines} from different fields (e.g., vehicles, robots) are indispensable to facilitate and automate tedious, time-consuming or hazardous tasks.
As a result, there is a strong incentive to continually advance them away from manual control towards greater levels of autonomy.
This increasing autonomy enables several new application areas, such as mapping unknown environments in exploration missions \cite{hu2020voronoi}, navigation in search and rescue operations \cite{jorge2019survey}, parking surveillance \cite{sarkar2019intelligent}, or delivering tasks in manufacturing and warehousing \cite{fragapane2021planning}. 
However, in all these use cases, machines are susceptible to a variety of dynamic environmental factors (object motion \cite{lu2016robust}, lighting \cite{keller2020illumination}, temperature \cite{kuhn2020analysis}, weather \cite{zhang2023perception}, etc.).
These factors have a direct impact on how their sensors perceive the environment, which, in turn, affects their subsequent actions \cite{veres2019deep, kuutti2020survey, rahman2022secure}.
Hence, to ensure the safety of both individuals and the machines themselves, special attention must be paid to reliable and robust sensors.
Cameras are nowadays the predominant sensors to perceive the environment, and are therefore the subject of this study.
To guarantee a camera's intended functionality, autonomy demands for self-health-maintenance, i.e., the task of continuously monitoring the behavior of the system and executing automatic countermeasures in case of a detected misbehavior.

Previous studies (e.g., \cite{lu2010camera, shim2018gradient, shin2019camera, liu2020estimating, mehra2020reviewnet}) have approached this task by monitoring and optimizing image features linked to general image quality (like sharpness, noise or dynamic range).
To this end, various automatic image quality maintenance techniques have been developed and are now part of a standard camera's imaging pipeline (auto-focus, auto-exposure, auto-calibration, etc.).
However, these techniques are typically decoupled from the downstream vision application (e.g., environment mapping, object detection, or navigation) and hence may not reach optimal application performance.
This is particularly true if the system can trade off image quality for other vision application benefits.
Moreover, each application has its own requirements for what is considered an optimal image quality.

This work closes the gap by proposing a general self-health-maintenance framework that strives for optimal application performance. 
The framework combines three key components: a continuous image quality monitoring, knowledge about the requirements of a downstream vision application, and a camera control for precise image quality adjustments.
We demonstrate the working principle of our framework on the exemplary application of object detection (as a representative modern vision application of great importance in various fields), and focus on motion blur and noise as typical undesired image properties (see Fig.~\ref{fig:eyeCatcher}).
Our modular design favors interpretability, explainability, and testability of individual components compared to end-to-end approaches.
Without loss of generality, our study analyzes: 
time-varying effects influencing blur and noise quality parameters (since any time-invariant effects are usually subtracted by camera calibration),
and region-wise effects, thus allowing us to consider spatially-varying problems.

We make the following contributions:
\begin{itemize}
    \item We propose a general framework to approach camera self-health-maintenance by maximizing the performance of arbitrary downstream vision applications through continuous monitoring and adjustment of image quality.
    
    \item We demonstrate our framework running in real-time on a real-world ground vehicle for the application of object detection (e.g., ``car'', ``pedestrian'') affected by motion blur and noise. %

    \item We evaluate two customized machine-learning (ML) based image blur and noise estimators in an experimental study:
    we compare them to four traditional state-of-the-art estimators on three datasets,
    account for five isolated and two combined blur and noise root causes grounded in knowledge of camera physics,
    and propose a post-processing step to re-enable blur estimation in presence of high noise.

    \item Our experiments yield practical recommendations for the robustness of camera monitoring applications.%

    \item We provide the source code of our experiments and of all estimators: \url{https://github.com/MaikWischow/Camera-Condition-Monitoring}.
\end{itemize}

\ifclearsectionlook\cleardoublepage\fi \section{Related Work} \label{sec:relatedWork}
Our study is closely related to active vision \cite{aloimonos1988active} and adaptive camera regulation \cite{murino1996adaptive} in that there are two connected tasks: online \emph{estimation} of the current vision state and execution of an \emph{action} to improve some target criterion. 
In the estimation task, we estimate major properties of the camera system state by assessing the quality of the image data \emph{it produces} in terms of blur and noise. 
Subsequently, we define actions that can be carried out to control the camera, therefore influence image properties (we demonstrate this for motion blur and noise) and hence optimize the system's performance for a target application (object detection in this work).

Motion blur can be directly approached at a hardware level by involving, e.g., an accelerometer \cite{chandrasekhar2018motion} or a self-designed sensor \cite{hamamoto2001computational}, but it is typically managed by automatic exposure control through image processing. 
Most image-based algorithms represent optimal exposure selection as a control problem on image quality indicators like the intensity entropy \cite{lu2010camera, shin2019camera}, gradients \cite{shim2018gradient, shin2019camera} and histograms \cite{li2015auto, torres2015optimal}, or approach it learning-based \cite{onzon2021neural}.
Our work is most similar to \cite{shin2019camera} and \cite{onzon2021neural}, thus we use both as comparison baselines. 

The study of Shin et al. \cite{shin2019camera} is likewise motivated by challenging image effects such as motion blur or noise, which ``can be dramatically
alleviated by carefully adjusting camera exposure parameters''.  
To this end, the authors propose an exposure time and gain control that maximizes image entropy, gradient strength, and gradient uniformity while minimizing noise.
However, this procedure does not account for the downstream vision applications that determine the required image quality (as most traditional approaches, such as \cite{lu2010camera, torres2015optimal, shim2018gradient, kettelgerdes2021correlating}).
Furthermore, they assume simplified additive Gaussian noise.
In contrast, we consider an extensive image formation pipeline that models the most important noise sources close to the camera physics (which is experimentally supported as being more realistic \cite{xu2018real, anaya2018renoir}).
We also include object detection performance as a feedback signal and hence aim for optimal downstream application performance.

Onzon et al. \cite{onzon2021neural} propose a neural network for auto-exposure control that is trained jointly, end-to-end with an object detector.
Unlike \cite{shin2019camera}, the authors consider an extensive noise formation pipeline.
On top of this, we assume a more comprehensive and realistic image formation by additionally including motion and defocus blur as well as all image corruptions occurring simultaneously and influencing each other (cf. \cite{tai2012motion}).
Moreover, we do not rely on a tailored end-to-end learning approach, but instead propose a more modular, extensible and interpretable concept: 
given the image data, we empirically determine a performance profile (adapted to the application) in terms of data quality metrics. 
To this end, we employ dedicated blur and noise estimators with real-time capabilities by adapting \cite{shin2019camera, shim2018gradient} and \cite{onzon2021neural} to focus on regions of interest.
The details are presented in the upcoming section.

Recent work also investigates environmental impacts (e.g., thermo-mechanical stress or mechanical vibrations) on defocus blur, resolving power, or camera calibration \cite{kettelgerdes2021correlating, kuhn2020analysis}. 
While these are based on fundamental studies, we focus on approaches directly applicable in practice.
\ifclearsectionlook\cleardoublepage\fi \section{Proposed System}
\label{sec:overview}
We introduce the proposed system in a top-down approach.
First, an overview is provided (Sec.~\ref{subsec:overview}).
Next, we explain its underlying working principle, which connects camera physics, blur, noise, and object detection performance (Sec.~\ref{sec:objectDetectionByTradingOffBlurAndNoise}).
We then detail the framework's components. 
We start with the creation of an application performance profile using the example of object detection (Sec.~\ref{sec:imgQualityAssessmentPerformanceProfiles}).
Finally, we present employed traditional and learning-based (ML) blur and noise estimation methods to quantify image quality objectively, and address the necessary changes we made (Secs.~\ref{sec:imageQ:blurestim} and \ref{sec:imageQ:noiseestim}). 

\subsection{Overview}
\label{subsec:overview}
\begin{figure}[t]
\centering
\includegraphics[width=\columnwidth]{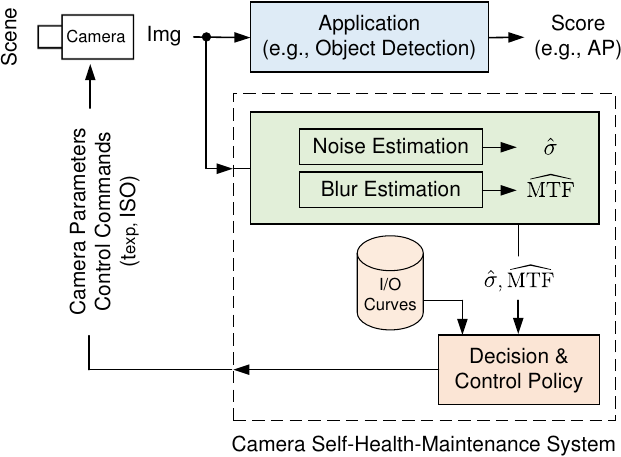}
\caption{\emph{System overview}.
The camera is constantly monitored by analyzing image corruptions (e.g., blur and noise). 
According to the estimated severity of such corruptions, camera control parameters (e.g., exposure time $t_\text{exp}$ and ISO gain) are recalculated to maximize application performance using the (offline determined) input/output (I/O) performance curves.
}
\label{fig:systemOverviewTesting}
\end{figure}

\begin{figure}[t]
\centering
\includegraphics[width=0.49\textwidth, clip]{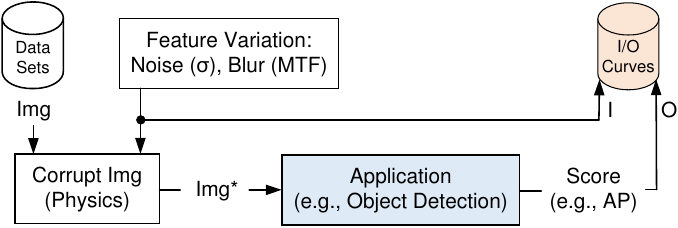}
\caption{\emph{Training the system (system identification)}. 
An offline sensitivity analysis determines the impact of physical image corruptions (e.g., blur and noise) on the performance
of a target application (e.g., object detection), 
and stores the results in input/output (I/O) performance curves. 
As input, image data close to the application domain are used.
}
\label{fig:systemOverviewTraining}
\end{figure}

Our proposed camera self-health-maintenance system consists of online testing (Fig.~\ref{fig:systemOverviewTesting}) and offline training parts (Fig.~\ref{fig:systemOverviewTraining}).

\begin{figure*}[t]
\centering
\includegraphics[width=\textwidth, trim=0 0 0 0, clip]{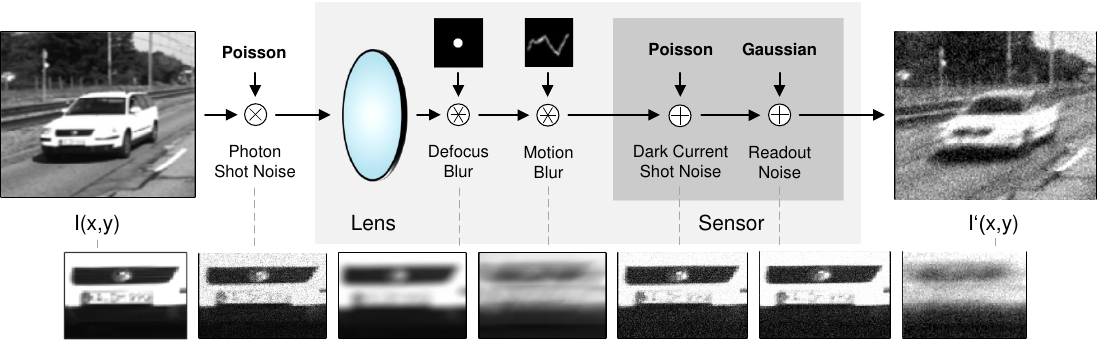}
\caption{\emph{Image formation process} of the considered camera system, including blur and noise models. 
A clean image $I(x,y)$ undergoes several physical processes that produce noise and blur, yielding the corrupted image $I'(x,y)$ (clean image patch vs. distinct corruptions in stated order). 
Noise is either signal-dependent or signal-independent, while blur is modelled as a convolution with a point spread function (PSF).
Details are provided in the supplementary material.}
\label{fig:modelBlurNoise}
\end{figure*}

Let us briefly introduce the offline training procedure first,
as training happens before the testing/inference phase.
We start with image datasets from a target application domain as input (e.g., object detection) and corrupt them according to an image formation pipeline (Fig.~\ref{fig:modelBlurNoise}).
The pipeline contains the most common (physics-based) sources of blur and noise affecting the camera condition, with realistic severity levels (Sec.~\ref{sec:experim:datasets}).
We quantify these levels using objective noise and blur metrics: noise level $\sigma$ and modulation transfer function (MTF) values, respectively.
Afterwards, we let our system's target application (object detection) evaluate these corrupted images.
We likewise quantify this performance in terms of the well known average precision score (AP, Sec.~\ref{sec:imgQualityAssessmentPerformanceProfiles}).  
Knowing each applied image corruption and the corresponding calculated application performance, the respective tuples are aggregated into input/output performance curves (IOPC), which is the final product of this training procedure.

The testing part (Fig.~\ref{fig:systemOverviewTesting}) has access to these IOPCs and analyzes each captured (yet unprocessed) camera image online using ML-based, real-time capable noise level and MTF estimators (Secs.~\ref{sec:imageQ:blurestim} and \ref{sec:imageQ:noiseestim}).
We evaluate their estimation and runtime performances compared to established state-of-the-art estimators (Sec.~\ref{sec:experiments}) for isolated and combined corruption cases, and propose a simple approach to improve blur estimation in case of interfering high noise levels (Sec.~\ref{sec:combinedEstimationResults}). %
If the estimated image quality does not meet the requirements for optimal application performance recorded in an IOPC, a control policy decides how to adjust camera parameters as countermeasure.
We propose two exemplary control policies using exposure time and ISO gain to trade off blur and noise.
They exploit the fact that object detectors are typically more sensitive to blur than to noise (Sec.~\ref{sec:sampleApplication}). 


\subsection{Optimize Object Detection by Trading off Blur and Noise}
\label{sec:objectDetectionByTradingOffBlurAndNoise}%
We now demonstrate how one can use the online blur/ noise estimators and the offline empirical IOPCs to control image quality and hence optimize object detection performance (Fig.~\ref{fig:systemOverviewTesting}).
Here we focus on actions tackling linear motion blur (Lin.~MB) 
because object detectors are substantially more sensitive to Lin.~MB than to noise (Fig.~\ref{fig:applicationMotionBlurPerformanceCurveResults}), 
and there is abundant motion blur in standard datasets like Udacity (Fig.~\ref{fig:udacity:challenging}).

We make the following considerations knowing the camera's physical processes. 
The main controllable influencing factor of motion blur is the camera's exposure time $t_\text{exp}$. 
We exploit the relations
\begin{equation}\label{eq:exposureTimeInfluence}
    \begin{gathered}
        t_\text{exp} \propto I \text{ and } t_\text{exp} \propto \text{MB} \sim \text{MTF}^{-1} \sim \text{AP},\\
        \text{ISO} \propto I \text{ and } \text{ISO} \propto \sigma \sim \text{AP}^{-1},
    \end{gathered}
\end{equation}
where AP is the average precision of the object detector.

Changing $t_\text{exp}$ by a factor $\alpha \doteq t^{\text{old}}_\text{exp} / t^{\text{new}}_\text{exp}$ equally changes the aggregated amount of light intensity $I$ and also the motion blur y the same factor (for simplicity and without loss of generality, we assume  linearity of the sensor digitization process \cite{wang2018development}).
To compensate for the changed light, we may alter the camera ISO gain by factor $\alpha$, which likewise changes the noise level~$\sigma$.
This relationship depends on the camera sensor architecture and whether the analog or digital signal is amplified \cite{igual2019photographic}.
We assume digital amplification as the worst case and thus a linear relation.
Hence, we can model the problem as an optimization one, i.e., determining $\alpha$ from the IOPCs to maximize the object detector's score:  
\begin{equation}\label{eq:APOptimization}
\alpha^\star = \argmax_{\alpha} ~ \text{AP}(\alpha \, \hat{\sigma}, \alpha \, \text{MB}(\widehat{\text{MTF}})),
\end{equation}
where $\widehat{\text{MTF}}$ and $\hat{\sigma}$ denote the online blur and noise estimations, respectively.
Note that $t_\text{exp} \propto \sigma_\text{DCSN}^2$ during optimization \cite[p.~3]{noiseSimulation}.
We drop this influence here for simplicity, since the small DCSN has no significant effect on $\hat{\sigma}$ in our setting.

\subsection{Empirical Input-Output Performance Curves}
\label{sec:imgQualityAssessmentPerformanceProfiles}
\begin{figure}[t]
\centering
\includegraphics[width=0.48\textwidth, clip]{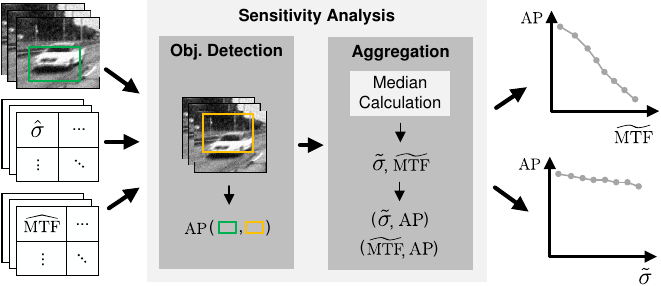}
\caption{\emph{Sensitivity analysis} of object detector performance for blur and noise. 
The detectors are evaluated on corrupted images resulting in average precision (AP) scores. 
For the true detection areas, corresponding patch-wise noise ($\hat{\sigma}$) and blur ($\widehat{\text{MTF}}$) estimations are aggregated to medians ($\tilde{\sigma}$ and $\widetilde{\text{MTF}}$) and, together with the APs, added to performance curves.}
\label{fig:objectDetection}
\end{figure}

Due to the non-linear and non-monotonic nature of vision applications, such as ML-based object detectors, we aim to determine system output sensitivities \emph{empirically} for different noise and blur levels (Fig.~\ref{fig:objectDetection}). 
In this work, we use YOLOv4~\cite{bochkovskiy2020yolov4} (YOLOv7 \cite{wang2023yolov7} in the supplementary material)
and Faster R-CNN~\cite{ren2015faster} as state-of-the-art real-time object detectors (with pre-trained models and default settings, applied on grayscale images).
The analysis is performed offline, but an online approach is also feasible. 

Let us explain the offline procedure on the example of images with a fixed blur level $\text{MTF}$ and noise level $\sigma$.
As input we assume $N_I$ images $\mathbf{I} \doteq \{I_{i}(x,y)\}^{N_I}_{i=1}$, where an image $I$ has $N_\text{GT}$ corresponding ground truth object detections ${\mathbf{B}_{I}^\text{GT} \doteq \{b^{\text{GT}}_{I, i}(x,y)\}^{N_\text{GT}}_{i=1}}$, patch-wise blur estimations $\widehat{\text{MTF}}_{I}(x,y)$, and noise estimations $\hat{\sigma}_{I}(x,y)$; $x$ and $y$ index respective pixel values. 
A pixel of a $b_{I}^{\text{GT}}(x,y)$ containing an object is defined as 1 and 0 otherwise.

First, both object detectors are applied to all images $I \in \mathbf{I}$ yielding the estimated object detections $\mathbf{B}_{I}^\text{D}$ per detector and image. 
Second, these $\mathbf{B}_{I}^\text{D}$ are evaluated against the $\mathbf{B}_{I}^\text{GT}$ using the well-known average precision (AP) metric, which we calculate following \cite{cartucho2018robust}.
In a subsequent aggregation step, we determine median blur and noise estimations of all image patches overlapping with the ground truth object detections, where
\begin{equation}
\tilde{\sigma} \doteq{} \text{med}( \{\hat{\sigma}_{I}(x,y)  |  b^{\text{GT}}_{I}(x,y) = 1, \forall I \in \mathbf{I}, \forall x,y \in \mathbb{N} \})
\end{equation}
and $\widetilde{\text{MTF}}$ is defined analogously.
To bound the complexity, we quantize the estimation parameter spaces into bins with $\tilde{\sigma} \in \{0, 5, \dots, 25\} \, \si{\DN}$ and $\widetilde{\text{MTF}} \in \{0.1, 0.2, \dots, 1.0\}$.
Finally, the resulting input-output tuples $(\tilde{\sigma}, \text{AP})$, $(\widetilde{\text{MTF}}, \text{AP})$ or $(\tilde{\sigma}, \widetilde{\text{MTF}}, \text{AP})$ are collected as performance curves (IOPCs).

\subsection{Blur Estimation (via the MTF)}
\label{sec:imageQ:blurestim}
\begin{figure}[t]
\centering
\includegraphics[width=0.48\textwidth, clip]{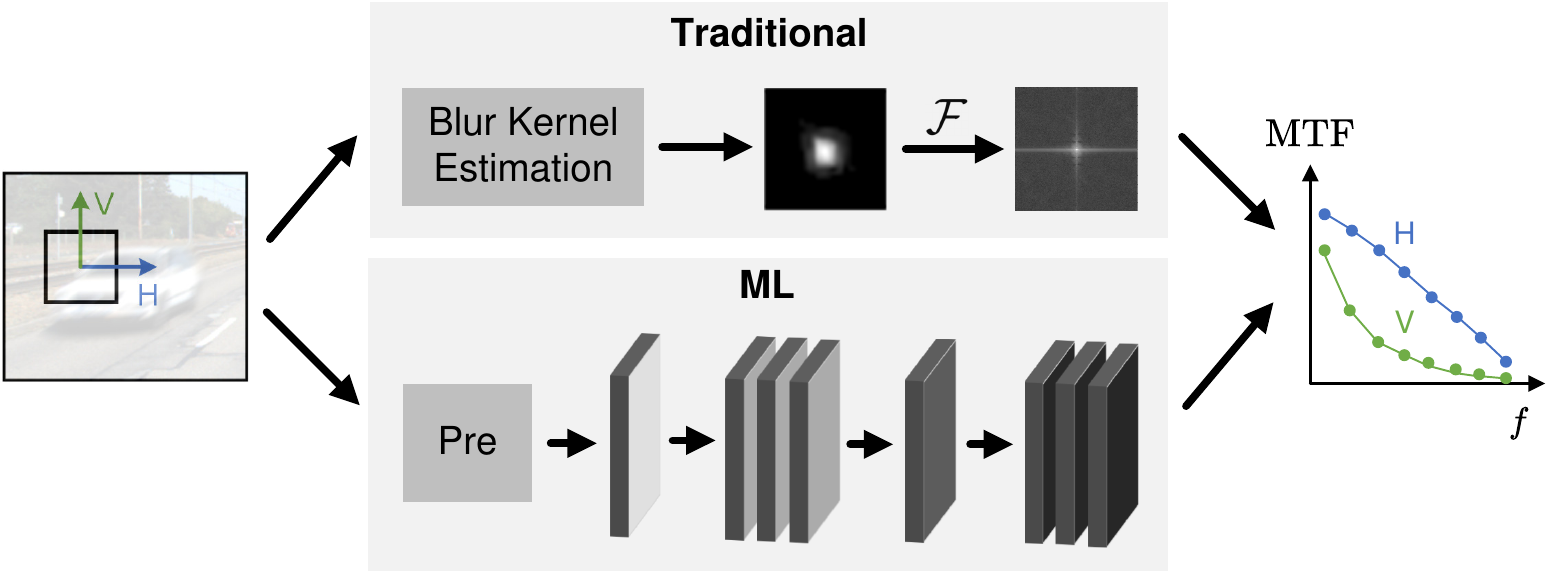}
\caption{\emph{Blur estimation} of traditional (top branch) and learning-based (bottom branch, ML) approaches. 
All methods input one or more image patches and output estimated MTF samples for pre-defined image frequencies ($f$) in the {\color{visioBlue} horizontal} ({\color{visioBlue} H}) and {\color{visioGreen} vertical} ({\color{visioGreen} V}) directions. 
Traditional methods first estimate a blur kernel, transform it into the Fourier space $\mathcal{F}$, and sample MTF values.
The learning-based method consists of a pre-processing stage (Pre) followed by a multi-layer CNN. 
}
\label{fig:blurEstimation}
\end{figure}

The goal of our image blur estimators is to predict the MTF given a possibly blurred input image patch $I^*$, where $I^*$ is assumed to be monochrome (i.e., grayscale) and of size $192 \times 192$ pixels (following the ML approach). Figure~\ref{fig:blurEstimation} summarizes the steps of the two main approaches.

\subsubsection{Traditional methods (non-learning--based)}
We use two baseline methods: 
``graph-based'' \cite{bai2018graph} (\underline{G}raph-\underline{B}ased \underline{B}lind Image Deblurring, or simply GBB) and ``simple local minimal intensity prior'' \cite{wen2020simple} 
(Patch-wise Minimal Pixels -- PMP) as traditional blur kernel estimators (top branch in Fig.~\ref{fig:blurEstimation}). 
Both estimators follow a maximum-a-posteriori framework 
\begin{equation}
\label{eq:MAPBlurKernel}
    \underset{I,h}{\min} ~ \mathcal{L}(I \circledast h, I^*) + \alpha G(I) + \beta R(h)
\end{equation}
to iteratively refine a clean latent image $I$ and the blur kernel~$h$
(see supplementary material for details on the image formation process).
The objective function~\eqref{eq:MAPBlurKernel} is the negative logarithm of the posterior distribution (thus maximization turns into minimization). 
It consists of a data fidelity term $\mathcal{L}$ that penalizes the deviations with respect to the observed image~$I^*$, and two regularizers $G$ and $R$ (prior knowledge) on the unknowns (with positive weights $\alpha,\beta$). 
The GBB approach represents images as graphs and employs a skeleton image with only strong gradients as a proxy for~$I$. 
It uses a re-weighted graph total variation prior $G(I)$ to favor bi-modal image histograms. 
The PMP method builds on top of the dark-channel prior, proposing a simplified patch-wise minimal pixel prior $G(I)$ that aims for sparse minimal pixel intensities with low computation complexity.
The resulting $h$ from each method is Fourier-transformed into the MTF and sampled at the same spatial frequencies as the learning-based approach (Fig.~\ref{fig:blurEstimation}), for better comparison.
We use the source code from \cite{bai2018graphGitHub,wen2020simpleGitHub}, setting the kernel size parameters to $31\times 31$ pixels.

\subsubsection{Learning-based Method}
We upgrade a learning-based approach \cite{cnnMtfEstimation} to \emph{directly} estimate MTF values from natural images (without estimating the kernel $h$ first). 
It consists of a pre-processing stage followed by a CNN.

The pre-processing stage includes four steps: 
($i$) Intensities are first scaled to $\interval{0}{1}$ and mean-normalized. 
($ii$) A rotation is applied to estimate the MTF in radial and tangential directions. 
($iii$) The Sobel-filtered image patch is passed as an additional channel to aid the MTF estimation procedure. 
($iv$) Channels are spatially down-sampled to enlarge the receptive field of early CNN layers.
We alter step ($ii$) to distinguish between estimations in horizontal and vertical directions to allow a comparison with the GBB and PMP baseline methods.

The CNN consists of a convolutional layer, seven residual blocks with strided convolutions, an intermediate feature representation layer, 
and three fully connected layers that regress the MTF outputs (bottom branch of Fig.~\ref{fig:blurEstimation}). 
The resulting output consists of eight MTF values in the range $\interval{0}{1} \smallSpace \si{\lines \per \px}$ at pre-defined spatial image frequencies. 

The training is supervised. 
In~\cite{cnnMtfEstimation}, pairs of sharp image patches and PSFs ($I,h$), synthetic or real, are collected. 
Their convolution leads to the training samples $I^*$; the respective MTF samples of the PSFs at the pre-defined frequencies serve as training labels. 
In contrast to~\cite{cnnMtfEstimation}, we blurred the sharp images by simulated random defocus and motion blur kernels (see Sec.~\ref{sec:experiments}), and retrained the CNN.
The original CNN weights are not publicly available and therefore cannot be used for comparison.

At inference time, we pass a batch of four input image patches, 
i.e., we stack temporally consecutive patches from the same sensor position, pre-process them independently, and input them into the CNN at once.
We expect better results this way according to \cite{cnnMtfEstimation}, although one patch works as well.
The obtained CNN output is then an (averaged) MTF estimation.

Since the original source code is not available, we re-implemented it with guidance from the authors.

\subsection{Noise Estimation}
\label{sec:imageQ:noiseestim}
\begin{figure}[t]
\centering
\includegraphics[width=0.35\textwidth, clip]{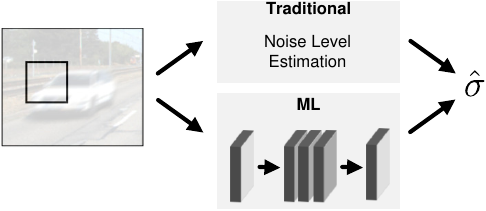}
\caption{
\emph{Noise estimation} of traditional or learning-based (ML, e.g., multi-layer CNN-based) approaches. 
Both approaches estimate a noise level $\hat{\sigma}$ for each input image patch.}
\label{fig:noiseEstimation}
\end{figure}

The goal of the image noise estimators is to predict the noise level $\sigma$ of a noise process given a noisy input image patch $\tilde{I}$, which is monochrome and of size of $128 \times 128$ pixels (following the ML approach).
Figure~\ref{fig:noiseEstimation} depicts the steps of the two main approaches.

\subsubsection{Traditional methods (non-learning--based)}
As baseline estimators we use the works of \cite{shin2005block} (self-implemented) and \cite{chen2015efficient} (with its code basis \cite{chen2015efficientGitHub}). Both are representatives of the two major noise estimation approaches in the literature: 

The adaptive Gaussian filtering method \cite{shin2005block} (B+F) uses the standard deviation of the most homogeneous image patches as a basis to calculate a Gaussian kernel that is used to filter such patches. 
The standard deviation of the difference between filtered and unfiltered patches leads to the estimated $\hat\sigma$. 
We increased the internal image patch size from $3 \times 16$ to $8 \times 16$ pixels as we observed better results on the selected datasets.

The method \cite{chen2015efficient} decomposes image patches via principal component analysis (PCA, also abbreviation of the method) into their eigenvalues and assigns the noise ratio to the smallest ones. 
In contrast to previous work, the authors tackle the problem of overestimating or underestimating noise theoretically and propose an efficient non-parametric algorithm for noise level estimation.

\subsubsection{Learning-based Method}
We use the work of \cite{tan2019pixelwise} as learning-based (ML) approach with its code basis \cite{tan2019pixelwiseGitHub}.
It was designed for pixel-wise noise level estimation from signal-dependent noisy images.
The noise model was assumed Gaussian with parameters accounting for photon and readout noise.

The CNN consists of 16 convolutional layers (including three residual blocks) 
and lacks pooling and interpolation layers, due to a known performance decrease for image noise tasks. 
The resulting output $\hat\sigma$ is estimated for each pixel, however, for a better comparison with baseline methods, we use the median over the patch as the noise level estimator.

The training in~\cite{tan2019pixelwise} is supervised and carried out by artificially adding noise with $\sigma\in\interval{0}{30} \, \si{\DN}$ to images from the Waterloo dataset \cite{ma2016waterloo}. 
We retrained the CNN in the same way using our noise model of Fig.~\ref{fig:modelBlurNoise} (details in Sec.~\ref{sec:experiments}).
Applying the original CNN weights to the noise model of \cite{noiseSimulation} and to real images failed; we could only reconstruct the authors' results for their simplified Gaussian noise process.
\ifclearsectionlook\cleardoublepage\fi \section{Experiments}
\label{sec:experiments}
We first describe the datasets used and the image corruptions applied (Sec.~\ref{sec:experim:datasets}). 
Subsequently, we evaluate the accuracy and runtime performances for the proposed blur and noise estimators separately (Secs.~\ref{sec:experimentsBlurEstimation} and \ref{sec:noiseEstimationResults}) and on combined image corruptions (Sec.~\ref{sec:combinedEstimationResults}).
All experiments are executed on an Intel Xeon W-2145 CPU and an NVIDIA Quadro RTX 6000 GPU, with the CNN methods running on the GPU.

\subsection{Datasets}
\label{sec:experim:datasets}
\begin{figure}[t]
\centering
\includegraphics[width=.9\columnwidth, trim=16 16 16 16, clip]{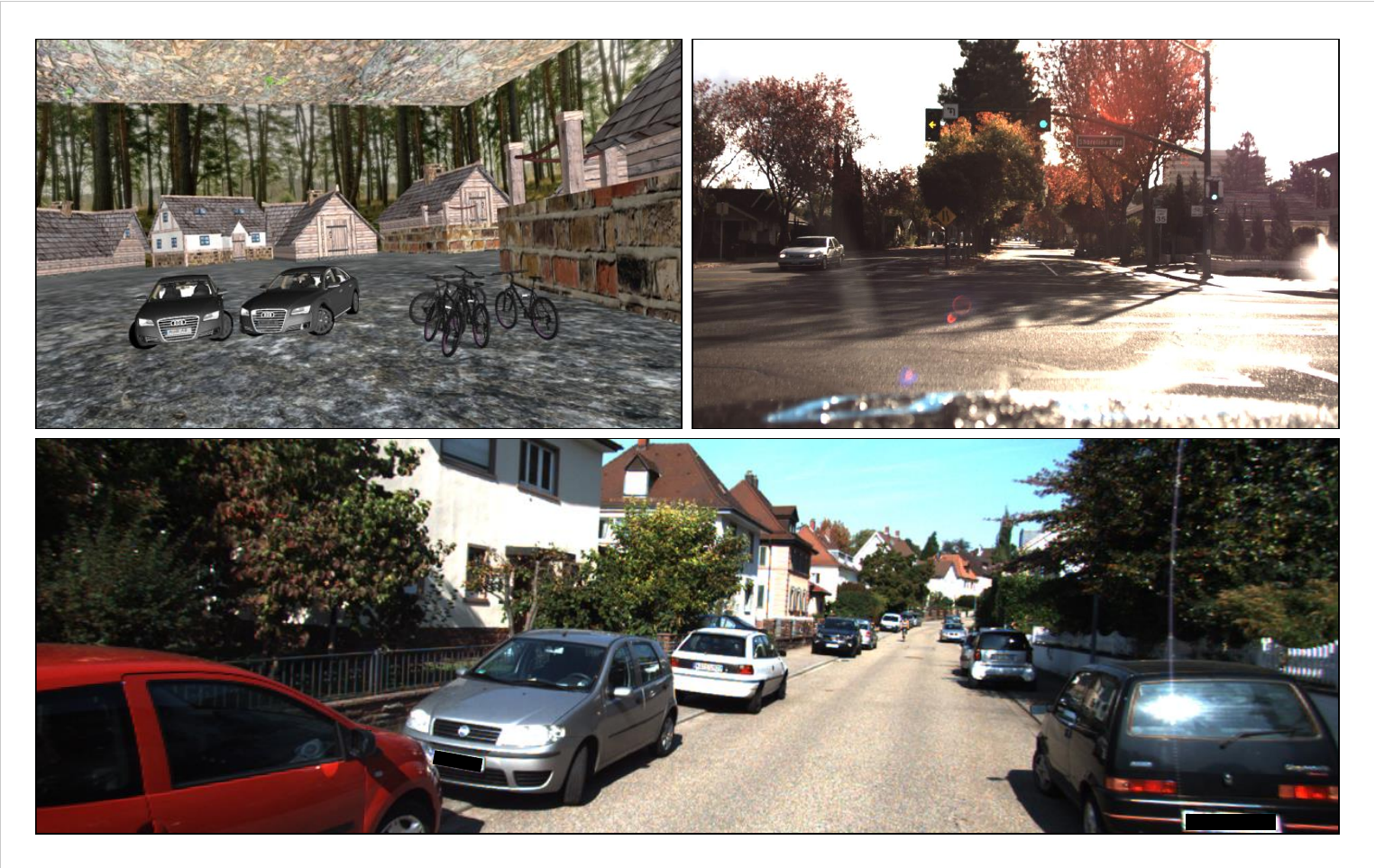}
\caption{\emph{Datasets}. 
Exemplary images from datasets \synth{} ($896 \times 768 \,\si{\px}$), 
Udacity ($1920 \times 1200 \,\si{\px}$) and KITTI ($1242 \times 375 \,\si{\px}$).
}
\label{fig:datasets}
\vspace{-1ex}
\end{figure}

We employ one simulated and two real-world datasets: \synth{}, KITTI~\cite{Geiger2012CVPR} and Udacity~\cite{udacity} (Fig.~\ref{fig:datasets}). 
We create \synth{} with the simulator \cite{irmisch2019simulation} to provide accurate ground truth for blur and noise estimation. 
\synth{} comprises \num{1000} images of a village environment acquired from different viewpoints and includes vehicles, such as cars and bikes. 
From KITTI we use the annotated object detection sub-dataset (with preceding frames), 
and from Udacity we use sub-dataset \#2.
We subsample KITTI and Udacity for two reasons: to reduce processing time and to remove (in all conscience) clearly visible blur/noise corrupted images that would bias estimation results (however, a residual risk of corruption in the natural images remains).
To this end, we pick \num{1000} images per dataset for noise estimation and \num{150} images for blur estimation, and match these numbers on \synth{}.
For blur, we only use image patches containing detected objects of interest.
Note that each dataset yields several image patches, depending on the image size and the blur/noise estimator used (e.g., 1000 Udacity images result in 135k patches for noise estimation).

We chose KITTI and Udacity as the most frequently used ones in the literature for real-world transportation scenarios that fit our requirements, providing a solid comparison baseline, and \synth{} to fully control the proposed blur and noise sources.
These datasets can be substituted with others that ideally contain minimal blur and noise, consecutive frames (for more stable median estimations, cf. Secs.~\ref{sec:noiseEstimationResults} and \ref{sec:noiseEstimationResults}), and annotated objects of interest.
Modern large-scale datasets (e.g., ROAD \cite{singh2022road}, Mapillary \cite{neuhold2017mapillary}, and Bdd100k \cite{yu2020bdd100k}) provide a greater variety of scenes and object classes, but typically consist of image sequences from different cameras having diverse geometric and radiometric statistics (e.g., their MTFs can differ significantly).
For an overview of more datasets (including KITTI and Udacity), we refer to \cite{yin2017use}.

All datasets are synthetically corrupted with controlled amounts of noise and blur using the models of Fig.~\ref{fig:modelBlurNoise}. 

\subsubsection*{Noise} 
Following the ``real noise'' studies in \cite{anaya2018renoir}, we generate noise with levels $\sigma \in \{5, 10, 15, 20, 25\} \,\si{\DN}$ (digital numbers on a $[0,255]$ scale). 
We apply default CMOS camera parameters from \cite{noiseSimulation} and study noise in isolation or in combination.
(\emph{i}) For isolated dark current shot noise (DCSN) and readout noise studies, we set the temperature to $T = \SI{330}{\K}$ and the exposure time to $t_{\text{exp}} = \SI{0.1}{\s}$.
(\emph{ii}) For the combined noise case we include \emph{all} noise sources, with random $T \in \SIrange{300}{330}{\K}$ and $t_{\text{exp}} \in \SIrange{0.002}{1}{\s}$ to emphasize different noise components in each image. 
In order to reach the desired $\sigma$, we amplify the (raw) noise in both settings. 

\subsubsection*{Blur} 
We synthesize blur kernels of size $d \in \{3, 7, 11, 15, 21\} \,\si{\px}$.
$d$ is the diameter for defocus kernels or the approximate path length for motion blur kernels.
Defocus blur kernels are calculated analytically (see supplementary material). 
Motion blur kernels are generated using \cite{githubMotionBlurKernelCode}, distinguishing between linear motion kernels (motion intensity parameter set to 0) and non-linear ones (parameter set to 1.0), and manually selecting the kernels that satisfy the target~$d$. 
We mitigate the influence of motion blur direction by evaluating four rotated versions of each kernel separately (rotating them 0, 45, 90, and 135 deg counterclockwise) and using their average estimation error as final result.%

We further propose two use cases for \emph{combined blur and noise} occurrences: 
(\emph{i}) \emph{Defocus blur and DCSN} (Defocus + DCSN) that might arise at high temperatures (as caused by direct Sun illumination) and with defocus induced by material stress in the optics setup \cite{kuhn2020analysis, kettelgerdes2021correlating}, and 
(\emph{ii}) \emph{photon noise and motion blur} (Photon + Motion) due to high exposure times and signal amplification, typical of low light conditions.   
\begin{figure*}[t]
    \noindent\begin{minipage}{.75\textwidth}
        \begingroup
            \pgfplotsset{every axis/.style={scale=0.55}}
            \pgfplotsset{cycle list/Set1, cycle multiindex* list={
                    mark list*\nextlist
                    Set1\nextlist
                }
            }
             \begin{tikzpicture}
 \scriptsize
   \begin{axis}[
        name=axis1,
        legend cell align={left},
        legend style={
          nodes={scale=0.65, transform shape},
          fill opacity=0.5,
          draw opacity=1,
          text opacity=1,
          at={(0.03,0.03)},
          anchor=south west,
          draw=white!80!black,
        },
        tick pos=left,
        xmajorgrids,
        xmin=0, xmax=0.6,
        xtick style={color=black},
        xtick={0, 0.1, 0.2, 0.3, 0.4, 0.5, 0.6},
        xticklabels={0, 0.1, 0.2, 0.3, 0.4, 0.5, 0.6},
        xlabel={Frequency [lines/px]},
        ymajorgrids,
        ymin=0, ymax=1.0,
        ytick style={color=black},
        ytick={0.2, 0.4, 0.6, 0.8, 1.0},
        yticklabels={0.2, 0.4, 0.6, 0.8, 1.0},
        title={\textbf{Sim}},
        ylabel={MTF},
        ylabel near ticks
    ]
        \addplot[color=black,line width=0.7mm,dotted] plot coordinates {
        (0.05,1.0)(0.1,0.99)(0.15,0.97)(0.2,0.95)(0.3,0.89)(0.4,0.81) (0.5,0.72) (0.6,0.62)
        };
        \addlegendentry{GT};
        \pgfplotsset{cycle list shift=-1}
        \addplot plot coordinates {
        (0.05,0.998) (0.1,0.989) (0.15,0.978) (0.2,0.959) (0.3,0.906) (0.4,0.817) (0.5,0.725) (0.6,0.591) 
        };
        \addlegendentry{CNN};
        \addplot plot coordinates {
        (0.05,0.832) (0.1,0.556) (0.15,0.504) (0.2,0.45) (0.3,0.324) (0.4,0.212) (0.5,0.145) (0.6,0.092) 
        };
        \addlegendentry{PMP};
        \addplot plot coordinates {
        (0.05,0.88) (0.1,0.666) (0.15,0.599) (0.2,0.544) (0.3,0.484) (0.4,0.385) (0.5,0.306) (0.6,0.223)
        };
        \addlegendentry{GBB};
        \addplot [forget plot, draw=none, name path=CNN_min] coordinates {
        (0.05,0.997) (0.1,0.986) (0.15,0.974) (0.2,0.95) (0.3,0.896) (0.4,0.806) (0.5,0.717) (0.6,0.582) 
        };
        \addplot [forget plot, draw=none, name path=CNN_max] coordinates {
        (0.05,0.998) (0.1,0.992) (0.15,0.982) (0.2,0.966) (0.3,0.914) (0.4,0.827) (0.5,0.733) (0.6,0.599) 
        };
        \addplot[red!40, opacity=0.2] fill between[of=CNN_min and CNN_max];
        \addplot [forget plot, draw=none, name path=PMP_min] coordinates {
         (0.05,0.568) (0.1,0.122) (0.15,0.113) (0.2,0.052) (0.3,0.024) (0.4,0.02) (0.5,0.012) (0.6,0.011) 
        };
        \addplot [forget plot, draw=none, name path=PMP_max] coordinates {
        (0.05,0.992) (0.1,0.968) (0.15,0.943) (0.2,0.894) (0.3,0.716) (0.4,0.499) (0.5,0.333) (0.6,0.213)
        };
        \addplot[blue!40, opacity=0.2] fill between[of=PMP_min and PMP_max];
        \addplot [forget plot, draw=none, name path=GBB_min] coordinates {
        (0.05,0.437) (0.1,0.035) (0.15,0.063) (0.2,0.119) (0.3,0.007) (0.4,0.014) (0.5,0.008) (0.6,0.011)
        };
        \addplot [forget plot, draw=none, name path=GBB_max] coordinates {
        (0.05,0.997) (0.1,0.987) (0.15,0.977) (0.2,0.958) (0.3,0.918) (0.4,0.863) (0.5,0.82) (0.6,0.654) 
        };
        \addplot[green!40, opacity=0.2] fill between[of=GBB_min and GBB_max];
  \end{axis}
  
  \begin{axis}[
        name=axis2,
        at=(axis1.right of south east),
        xshift=+3mm,
        tick pos=left,
        xmajorgrids,
        xmin=0, xmax=0.6,
        xtick style={color=black},
        xtick={0, 0.1, 0.2, 0.3, 0.4, 0.5, 0.6},
        xticklabels={0, 0.1, 0.2, 0.3, 0.4, 0.5, 0.6},
        ymajorgrids,
        ymin=0, ymax=1.0,
        ytick style={color=black},
        ytick={0.2, 0.4, 0.6, 0.8, 1.0},
        yticklabels=\empty,
        title={\textbf{KITTI}}
    ]
        \addplot plot coordinates {
        (0.05,0.998) (0.1,0.99) (0.15,0.979) (0.2,0.96) (0.3,0.906) (0.4,0.815) (0.5,0.727) (0.6,0.59) 
        };
        \addplot plot coordinates {
        (0.05,0.964) (0.1,0.858) (0.15,0.75) (0.2,0.598) (0.3,0.409) (0.4,0.249) (0.5,0.168) (0.6,0.107)  
        };
        \addplot plot coordinates {
        (0.05,0.992) (0.1,0.966) (0.15,0.944) (0.2,0.905) (0.3,0.817) (0.4,0.696) (0.5,0.581) (0.6,0.443)
        };
        \addplot [forget plot, draw=none, name path=CNN_min] coordinates {
        (0.05,0.996) (0.1,0.985) (0.15,0.969) (0.2,0.946) (0.3,0.888) (0.4,0.795) (0.5,0.707) (0.6,0.576) 
        };
        \addplot [forget plot, draw=none, name path=CNN_max] coordinates {
        (0.05,0.999) (0.1,0.995) (0.15,0.987) (0.2,0.971) (0.3,0.924) (0.4,0.832) (0.5,0.746) (0.6,0.602)
        };
        \addplot[red!40, opacity=0.2] fill between[of=CNN_min and CNN_max];
        \addplot [forget plot, draw=none, name path=PMP_min] coordinates {
         (0.05,0.878) (0.1,0.57) (0.15,0.343) (0.2,0.143) (0.3,0.069) (0.4,0.039) (0.5,0.021) (0.6,0.026) 
        };
        \addplot [forget plot, draw=none, name path=PMP_max] coordinates {
        (0.05,0.994) (0.1,0.975) (0.15,0.957) (0.2,0.92) (0.3,0.776) (0.4,0.486) (0.5,0.329) (0.6,0.206)
        };
        \addplot[blue!40, opacity=0.2] fill between[of=PMP_min and PMP_max];
        \addplot [forget plot, draw=none, name path=GBB_min] coordinates {
        (0.05,0.971) (0.1,0.895) (0.15,0.854) (0.2,0.779) (0.3,0.641) (0.4,0.443) (0.5,0.287) (0.6,0.119) 
        };
        \addplot [forget plot, draw=none, name path=GBB_max] coordinates {
        (0.05,0.997) (0.1,0.99) (0.15,0.982) (0.2,0.967) (0.3,0.932) (0.4,0.879) (0.5,0.83) (0.6,0.695) 
        };
        \addplot[green!40, opacity=0.2] fill between[of=GBB_min and GBB_max];
  \end{axis}
  
  \begin{axis}[
        name=axis3,
        at=(axis2.right of south east),
        xshift=+6mm,
        tick pos=left,
        xmajorgrids,
        xmin=0, xmax=0.6,
        xtick style={color=black},
        xtick={0, 0.1, 0.2, 0.3, 0.4, 0.5, 0.6},
        xticklabels={0, 0.1, 0.2, 0.3, 0.4, 0.5, 0.6},
        ymajorgrids,
        ymin=0, ymax=1.0,
        xlabel near ticks,
        ytick style={color=black},
        ytick={0.2, 0.4, 0.6, 0.8, 1.0},
        yticklabels=\empty,
        title={\textbf{Udacity}}
    ]
        \addplot plot coordinates {
        (0.05,0.992) (0.1,0.964) (0.15,0.935) (0.2,0.877) (0.3,0.751) (0.4,0.56) (0.5,0.451) (0.6,0.313) 
        };
        \addplot plot coordinates {
        (0.05,0.921) (0.1,0.742) (0.15,0.616) (0.2,0.477) (0.3,0.312) (0.4,0.182) (0.5,0.119) (0.6,0.078) 
        };
        \addplot plot coordinates {
        (0.05,0.979) (0.1,0.921) (0.15,0.868) (0.2,0.777) (0.3,0.627) (0.4,0.475) (0.5,0.374) (0.6,0.253) 
        };
        \addplot [forget plot, draw=none, name path=CNN_min] coordinates {
        (0.05,0.979) (0.1,0.925) (0.15,0.86) (0.2,0.79) (0.3,0.65) (0.4,0.466) (0.5,0.315) (0.6,0.163)
        };
        \addplot [forget plot, draw=none, name path=CNN_max] coordinates {
        (0.05,0.999) (0.1,0.995) (0.15,0.987) (0.2,0.971) (0.3,0.932) (0.4,0.848) (0.5,0.76) (0.6,0.599)
        };
        \addplot[red!40, opacity=0.2] fill between[of=CNN_min and CNN_max];
        \addplot [forget plot, draw=none, name path=PMP_min] coordinates {
         (0.05,0.768) (0.1,0.38) (0.15,0.188) (0.2,0.11) (0.3,0.088) (0.4,0.028) (0.5,0.013) (0.6,0.024) 
        };
        \addplot [forget plot, draw=none, name path=PMP_max] coordinates {
        (0.05,0.976) (0.1,0.908) (0.15,0.842) (0.2,0.723) (0.3,0.506) (0.4,0.296) (0.5,0.209) (0.6,0.14)
        };
        \addplot[blue!40, opacity=0.2] fill between[of=PMP_min and PMP_max];
        \addplot [forget plot, draw=none, name path=GBB_min] coordinates {
        (0.05,0.948) (0.1,0.835) (0.15,0.772) (0.2,0.616) (0.3,0.458) (0.4,0.217) (0.5,0.099) (0.6,0.033) 
        };
        \addplot [forget plot, draw=none, name path=GBB_max] coordinates {
        (0.05,0.986) (0.1,0.946) (0.15,0.909) (0.2,0.841) (0.3,0.718) (0.4,0.591) (0.5,0.513) (0.6,0.447) 
        };
        \addplot[green!40, opacity=0.2] fill between[of=GBB_min and GBB_max];
  \end{axis}
\end{tikzpicture}
        \endgroup
    \end{minipage}
    \qquad
    \noindent\begin{minipage}{.15\textwidth}
        \small
        \begin{tabular}{lr}
             \toprule
             \multicolumn{2}{l}{Mean Runtime [s]}\\  
             \midrule
             $\text{CNN}_{\text{MTF}}$ &  0.24\\
             PMP &  12.69\\
             GBB &  13.07\\
             \midrule
             $\text{CNN}_{\sigma}$ &  0.002\\
             PCA & 0.002\\
             B+F & 0.005\\
             \bottomrule
        \end{tabular}
    \end{minipage}
    \caption{\emph{Blur estimation of uncorrupted datasets (i.e., ``ground truth'')}.
    Left: Median, minimum and maximum blur estimations of the uncorrupted datasets (depicted by sampled points with interpolation in between and the shaded areas, respectively; horizontal direction only). 
    Right: Mean runtime estimations per image patch (for $\text{CNN}_{\text{MTF}}$ per input batch of four images).}
    \label{fig:uncorruptedBlurEstimationsAndRuntime}
\end{figure*}
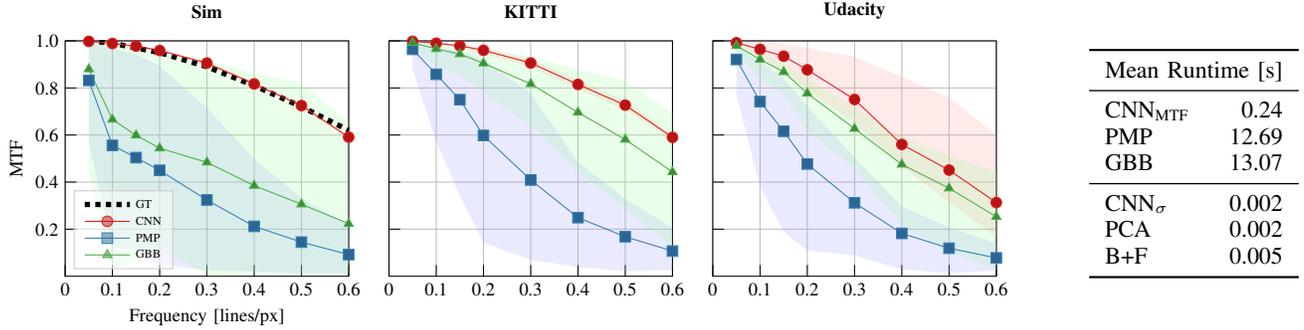
\ifclearsectionlook\cleardoublepage\fi 
\subsection{Blur Estimation}
\label{sec:experimentsBlurEstimation}%
\begin{table*}[t]
    \caption{\emph{Blur estimation of synthetically corrupted datasets}.
    Left: Ground truth blur kernels and average mean absolute errors (AMAE) of horizontal and vertical median blur estimations [\%]. The best results per kernel and dataset are highlighted in bold. 
    Right: Typical GBB/PMP kernel estimations with undesired artifacts (compare to respective ground truth kernels).}\label{tab:blurEstimationResults}
    \centering
    \begin{adjustbox}{max width=0.85\linewidth}
    \setlength{\tabcolsep}{4pt}
    \begin{tabular}{llccccccccccccccccc}
        \toprule
        &&\multicolumn{5}{c}{Defocus Blur}&\multicolumn{5}{c}{Linear Motion Blur}&\multicolumn{5}{c}{Non-linear Motion Blur}%
        \\
        \cmidrule(r){3-7}\cmidrule(r){8-12}\cmidrule(r){13-17}   
        Size [$\si{px}$] && 3 & 7 & 11 & 15 & 21%
        & 3 & 7 & 11 & 15 & 21%
        & 3 & 7 & 11 & 15 & 21%
        \\
        Kernel && \includegraphics[width=0.023\textwidth]{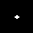}%
        & \includegraphics[width=0.023\textwidth]{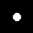}%
        & \includegraphics[width=0.023\textwidth]{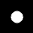}%
        & \includegraphics[width=0.023\textwidth]{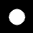}%
        & \includegraphics[width=0.023\textwidth]{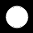}%
        & \includegraphics[width=0.023\textwidth]{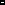}%
        & \includegraphics[width=0.023\textwidth]{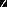}%
        & \includegraphics[width=0.023\textwidth]{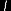}%
        & \includegraphics[width=0.023\textwidth]{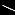}%
        & \includegraphics[width=0.023\textwidth]{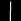}
        & \includegraphics[width=0.023\textwidth]{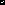}%
        & \includegraphics[width=0.023\textwidth]{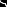}%
        & \includegraphics[width=0.023\textwidth]{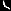}%
        & \includegraphics[width=0.023\textwidth]{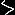}%
        & \includegraphics[width=0.023\textwidth]{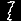}\\
        \midrule
        \small{Sim} &
            CNN%
                    & \textbf{0.7} & \textbf{1.8} & \textbf{2.1} & \textbf{0.5} & \textbf{1.1}%
                    & \textbf{6.3} & \textbf{10.3} & 9.4 & 9.4 & \textbf{7.7}%
                    & \textbf{2.9} & 12.2 & 11.4 & 19.5 & 25.0\\
            &PMP%
                    & 13.9 & 5.2 & 3.0 & 5.3 & 6.7%
                    & 37.2 & 17.8 & 13.3 & \textbf{7.3} & 16.6%
                    & 21.8 & 14.0 & 13.8 & 9.8 & \textbf{11.5}\\
            &GBB%
                    & 2.7 & 3.8 & 6.3 & 8.4 & 17.6%
                    & 31.6 & 11.4 & \textbf{7.7} & \textbf{7.3} & 14.5%
                    & 17.7 & \textbf{8.3} & \textbf{8.2} & \textbf{9.7} & 15.0\\[0.9ex]
        \small{KITTI} &
            CNN%
                    & \textbf{0.3} & 5.3 & 2.7 & \textbf{2.3} & \textbf{0.7}%
                    & \textbf{3.4} & 10.9 & 9.9 & 9.1 & \textbf{6.6}%
                    & \textbf{4.0} & 14.1 & 11.2 & 14.1 & 9.2\\
            &PMP%
                    & 5.8 & \textbf{2.3} & \textbf{1.5} & 2.9 & 4.8%
                    & 37.2 & 12.2 & 8.7 & \textbf{4.1} & 9.5%
                    & 22.5 & 7.9 & 7.3 & 5.2 & 4.2\\
            &GBB%
                    & 3.2 & 2.9 & 2.6 & \textbf{2.3} & 9.3%
                    & 13.4 & \textbf{5.8} & \textbf{5.3} & 4.7 & 7.1%
                    & 8.5 & \textbf{4.0} & \textbf{5.2} & \textbf{3.5 }& \textbf{3.7}\\[0.9ex]
                    
        \small{Udacity} &
            CNN%
                    & \textbf{2.7} & \textbf{0.9} & \textbf{0.6} & \textbf{0.3} & \textbf{1.4}%
                    & \textbf{16.2} & \textbf{10.8} & 9.8 & 11.4 & \textbf{7.9}%
                    & \textbf{9.6} & 10.3 & 11.9 & 16.0 & 19.2\\
            &PMP%
                    & 15.1 & 5.6 & 4.0 & 3.9 & 3.9%
                    & 34.3 & 14.2 & 11.5 & \textbf{8.1} & 12.4%
                    & 23.5 & 12.0 & \textbf{11.0} & \textbf{8.1} & \textbf{7.6}\\
            &GBB%
                    & 2.8 & 8.8 & 8.6 & 8.9 & 23.2%
                    & 21.7 & 12.2 & \textbf{8.5} & 12.2 & 21.1%
                    & 10.6 & \textbf{10.1} & 11.6 & 13.5 & 16.6\\
        \bottomrule
    \end{tabular}%
    \end{adjustbox}
    \qquad\quad
    \begin{minipage}{.16\textwidth}
         \includegraphics[width=1.00\textwidth]{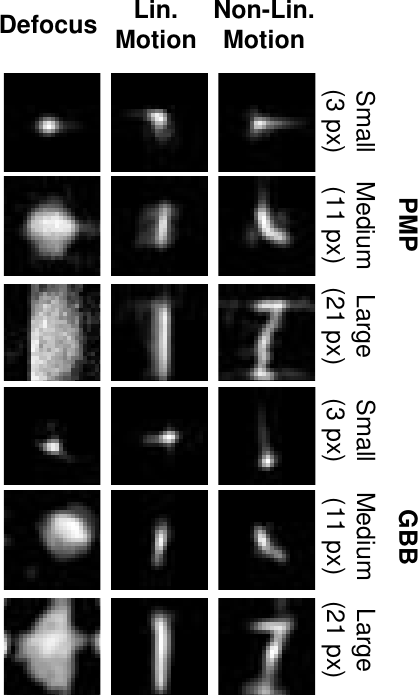}
    \end{minipage}
\end{table*}

We assess blur estimation accuracy in terms of the average mean absolute error (AMAE) between a robust MTF estimation ($\widehat{\text{MTF}}$, with $\approx\SI{5}{\%}$ outliers rejected) and ground truth (GT) samples at eight frequencies ($f_i$) each in horizontal (H) and vertical (V) image directions ($w$):
\begin{equation}
    \begin{split}
        \label{eq:amaedef}
        \text{AMAE} & \doteq \frac1{2} \sum_{w=\left\{\text{H,V}\right\}} \text{MAE}(w),\\
        \text{MAE}\,(w) & \doteq \frac1{8} \sum_{i \, = \, 1}^{8} \left| \text{MTF}^{\text{GT}}_{w}(f_i) - \widehat{\text{MTF}}_{w}(f_i) \right|.
    \end{split}
\end{equation}

\begin{figure}[t]
\centering
\includegraphics[width=0.8\columnwidth, trim= 16 16 16 16, clip]{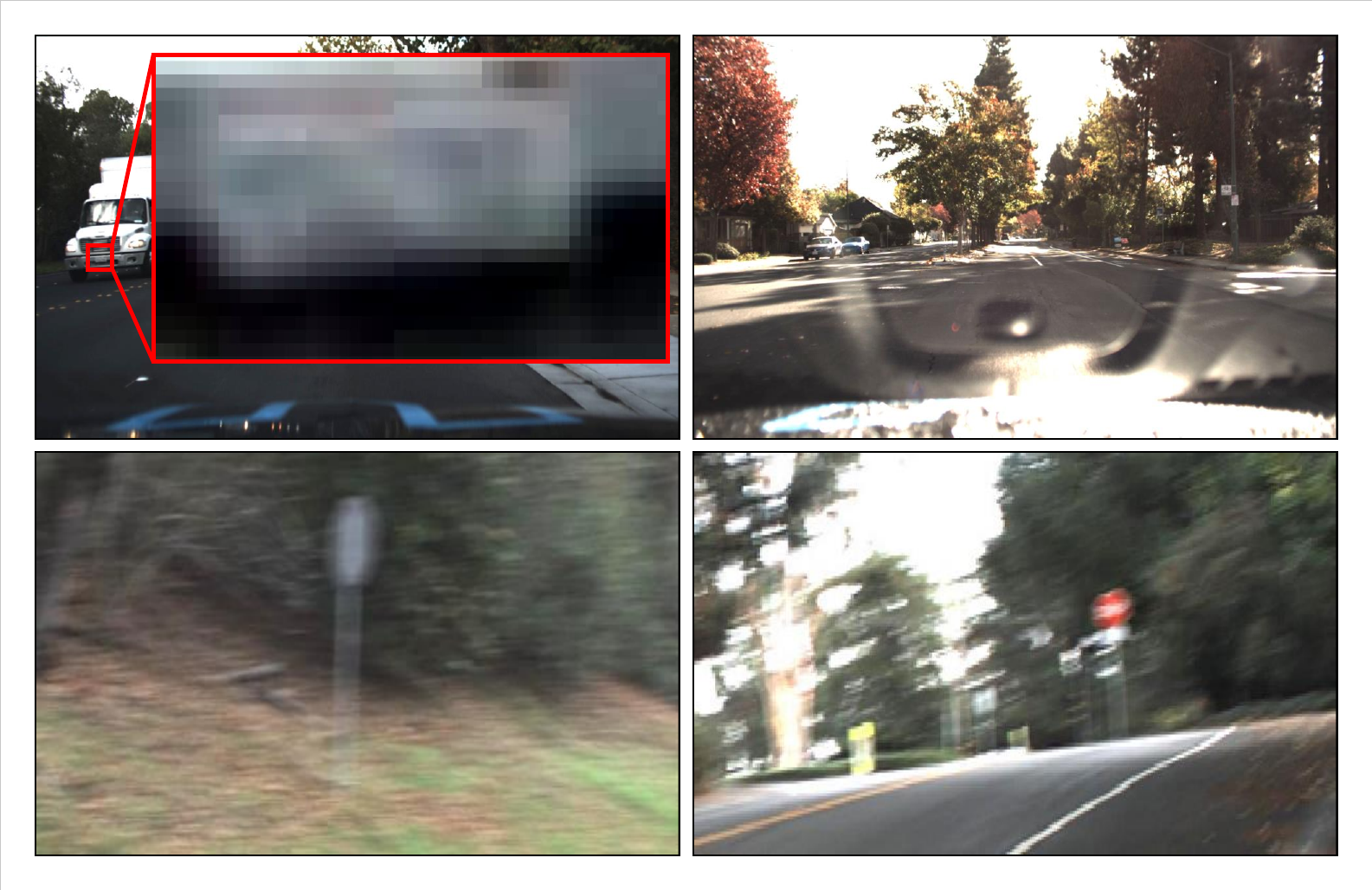}
\caption{\emph{Challenging conditions in the Udacity dataset}. 
From top left: Slight motion blur ($\SI{3}{\px}$) in moving direction, light reflections and two examples of severe motion blur.}
\label{fig:udacity:challenging}
\end{figure}

We first calculated (robust) median, minimum and maximum estimations for the uncorrupted datasets in Fig.~\ref{fig:uncorruptedBlurEstimationsAndRuntime}. 
In the \synth{} case, we could determine $\text{MTF}^\text{GT}$ by evaluating a Siemens Star (generated in the simulator) with the tool~\cite{meissner2021determination}. 
For the real-world datasets however, there are no known GT values, but we expect similar sharp images and hence we plot the estimations for comparison. 

Analyzing Fig.~\ref{fig:uncorruptedBlurEstimationsAndRuntime} we make five major observations: 
($i$) The CNN estimates a nearly ideal MTF with hardly any variance in the \synth{} case and provides similarly confident estimations for KITTI.
($ii$) Contrary to expectations, the CNN estimates a more uncertain and lower MTF for Udacity. 
Concerning this, we found challenging effects that influenced the estimation, like frequent windshield reflections and regular slight motion blur in the moving direction, despite our pre-selection of images. 
The traditional estimators (GBB/PMP) are also affected, producing lower median estimations than for KITTI. 
($iii$) The variances of GBB/PMP shrink from \synth{}, via KITTI towards Udacity. 
($iv$) GBB performs noticeably worse in \synth{}. 
We ascribe its low median and large variance to the lack of image gradient diversity of the \synth{} dataset (GBB relies on gradients, but strong horizontal edges are scarce in \synth{}). 
($v$) PMP produces generally low estimations and its maxima are far from the GT (\synth{}) or expected GT (real-world) values.

\begin{figure*}[t]
    \begingroup
        \pgfplotsset{every axis/.style={scale=0.435}}
        \pgfplotsset{cycle list/Set1, cycle multiindex* list={
                mark list*\nextlist
                Set1\nextlist
            },
        }
        \input{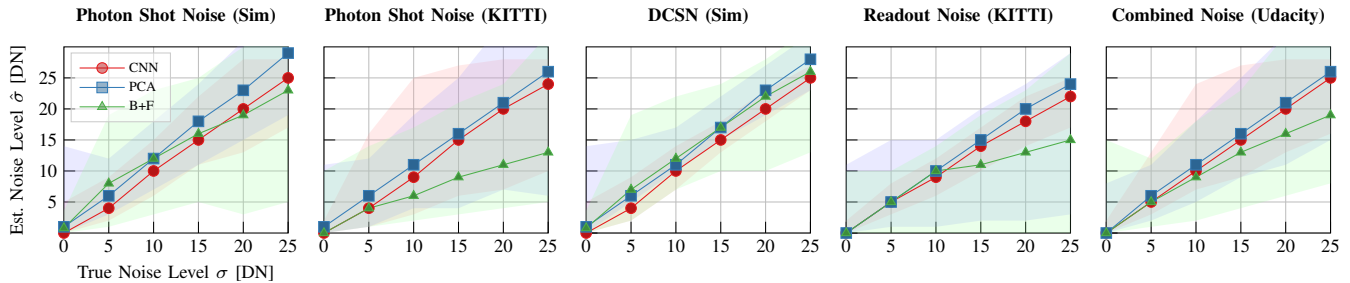}
    \endgroup
    \vspace{-1ex}
    \caption{\emph{Noise estimation of corrupted datasets.}
    Median, minimum and maximum statistics (depicted by sampled points with interpolation in between and the shaded areas, respectively) of the three proposed noise estimators (CNN, PCA, B+F) as the noise level $\sigma$ increases (from 0 to 25 grayscale levels, \si{\DN}), for several types of noise (Photon Shot, DCSN, etc.) and datasets (\synth{}, KITTI, Udacity). 
    The last plot shows the effect of combining all noise types (on the Udacity dataset).}
    \vspace{-1ex}
    \label{fig:noiseEstimationResults}
\end{figure*}

Next, we corrupted the datasets with the generated blur kernels and used the sampled MTFs of the kernels as ground truth. The blur AMAE scores are summarized in Table~\ref{tab:blurEstimationResults}.
We make the general observation that PMP and GBB ---unlike the CNN--- usually perform worst for small (\SI{3}{\px}) and large (\SI{21}{\px}) kernel sizes, respectively.
This often manifests in undesired artifacts like smear or cuttings in these estimations (see kernels in Tab.~\ref{tab:blurEstimationResults} right).
The decreased performance for small blur cases is in agreement with the results from Fig.~\ref{fig:uncorruptedBlurEstimationsAndRuntime}, where GBB and particularly PMP produce lower median estimations and higher variance for \synth{}/KITTI, and lower variance for the already corrupted Udacity. 
Since GBB/PMP follow a coarse-to-fine approach, more internal iterations would enhance the level of detail of the kernel and thus produce smaller errors (at the expense of computational cost). %
On the other hand, larger kernel estimations improve as larger image patches are used. 
The authors of GBB~\cite{bai2018graph} suggest kernels be much smaller than the image to have a well-defined blur estimation problem. 
We further regularly observe larger estimation errors for Udacity. %
This confirms that Udacity is already corrupted by blur and/or the estimations are influenced by challenging conditions (Fig.~\ref{fig:udacity:challenging}).

Apart from the already mentioned small/large kernels, all methods estimate defocus well (Tab.~\ref{tab:blurEstimationResults}). 
Nevertheless, the CNN delivers the most accurate results. 
GBB considers the common simplification of Gaussian blur for defocus, whereas PMP does not and tends to perform slightly better than GBB.

The CNN also estimates linear motion blur comparably well but (except for small/large kernels) GBB tends to produce the smallest errors. 

Non-linear motion estimation results (also in Tab.~\ref{tab:blurEstimationResults}) differ for the CNN method, 
which tends to produce larger errors towards the larger (and more complex) kernels compared to the traditional estimators and the linear case.
We interpret this as a larger uncertainty and conclude that the CNN might not be appropriate for estimation of complex non-linear motion kernels. 
In contrast, the scores of GBB/PMP are more accurate among the different kernels and datasets (with GBB a bit better).
This slightly better motion blur estimation performance of GBB compared to PMP is consistent with the experiments in~\cite{bai2018graph}, where PMP is compared to the work of Pan et al.~\cite{pan2016blind} that first proposes a dark channel prior.

\emph{Computational performance}:
In terms of runtime, we see from the table in Fig.~\ref{fig:uncorruptedBlurEstimationsAndRuntime} that the CNN executes more than $\times \num{50}$ faster than GBB/PMP and moves in the realm of real-time capability. 
We also found that CNN requires 98\% of its runtime for serial data pre-processing, which can be improved by vectorization.
Although the CNN itself executes on a GPU, the running times of current GBB/PMP implementations (running on the CPU) are too long to be practical for a condition monitoring application (especially for multiple image patches).

\emph{In summary}, the GBB and PMP methods are in general not accurate for blur-free or small/large blur kernel estimation on the image patch sizes used, and available implementations are not real-time capable.
Nevertheless, they provide the best estimates for medium-sized linear and non-linear motion blur kernels. 
The CNN method, on the other hand, might not be suited for complex non-linear motion kernels, but performs well in terms of defocus, linear motion and real-time requirements.
If non-linear motion blur can be circumvented (e.g., with short exposure times or slow motions), the CNN method can be employed for monitoring a camera's condition.

\ifclearsectionlook\cleardoublepage\fi 
\subsection{Noise Estimation} 
\label{sec:noiseEstimationResults}%
We evaluate the proposed noise estimators by comparing their robust median, minimum and maximum statistics (rejecting $\approx\SI{5}{\%}$ outliers) against the controlled ground truth noise levels.
Results are reported in Fig.~\ref{fig:noiseEstimationResults}.
Since we obtained comparable results for DCSN and readout noise per dataset,
we dropped similar plots.

\begin{figure*}[t]
    \begingroup
        \pgfplotsset{every axis/.style={scale=0.55}}
        \input{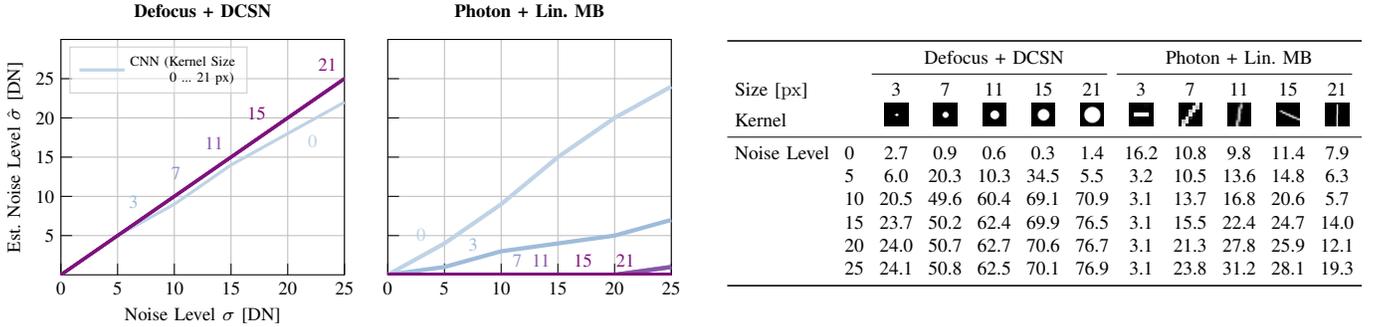}
    \endgroup
    \vspace{-1ex}
    \caption{\emph{Combined estimations of blur and noise} for two image corruption configurations: 
    Defocus + DCSN and Photon + Linear Motion Blur, both on the Udacity dataset.
    Plots of the median noise estimation (Left) 
    and table with median blur estimation (AMAE~\eqref{eq:amaedef} in \%) (Right)
    for different noise levels and kernel sizes. 
    Noise estimated for different blur kernel sizes is color coded from blue to purple. 
    However, differences are almost indistinguishable at this scale.
    }
    \vspace{-1.5ex}
    \label{fig:combinedEstimationResults}
\end{figure*}

We first observed that B+F and PCA methods are prone to structural misestimation: 
both over-estimate low noise levels, and B+F under-estimates high noise levels.
These phenomena have been already reported and are characteristic of the corresponding model family \cite{shin2005block,chen2015efficient}.
Moreover, all methods tend to strongly under-estimate noise in natural images, which even reduces the median performance of the B+F method. 
We observe this behavior in over-exposed areas where most pixels are in saturation, which is expected from vehicle camera images containing large sky areas. 
The CNN method is less vulnerable since it learned employing fewer meaningful pixels; \cite{shin2019camera} omits such image regions under the assumption that under-/over-saturated patches ``cannot contain noise'' (which only holds for \emph{completely} saturated regions).

Another observation is the striking difference between the signal-dependent and signal-independent noise cases. 
signal-dependent photon shot noise increases the variance of all estimators, especially on real-world data. 
We observed that large variations in bright and dark intensity areas within one image patch led to over- and under-estimation, respectively. 
The CNN noise level is limited here since it was trained with $\sigma \leq \SI{30}{\DN}$.
If all noise types occur simultaneously (last plot in Fig.~\ref{fig:noiseEstimationResults}) the estimations become more accurate and more robust than in the case of all noise being attributed to photon shot noise.
According to the observations of \cite{xu2018real, anaya2018renoir}, realistic noise follows a combined Poisson-Gaussian distribution, and the Poisson part is troublesome for the noise estimators (in particular for those with Gaussian assumptions). 
Hence, we consider isolated photon shot noise as the worst case scenario. 
The CNN and PCA methods perform similarly if signal-dependent photon shot noise is included, and the CNN is more reliable (smaller variance) otherwise. 
In terms of denoising, similar results have been shown by comparing traditional and learning-based methods on real data \cite{xu2018real}.

\emph{Computational performance}: 
Regarding runtime (Fig.~\ref{fig:uncorruptedBlurEstimationsAndRuntime}, right table), CNN and PCA executed fastest, with an average of \SI{2}{\ms} per patch, but in the same order of magnitude as the B+F (\SI{5}{\ms}). 
All noise estimators are real-time capable and considerably faster than blur estimators.

\emph{Summarizing}, the CNN and PCA methods are accurate in median 
but their reliability decreases the stronger the photon shot noise is. 
In case of signal-independent noise only, the CNN performs by far most reliably.
Since PCA is prone to structural misestimation (e.g., over-exposed areas, small noise levels), we suggest using the CNN for condition monitoring applications. 
Finally, the reliability of PCA and CNN could be improved by using the median estimation from consecutive frames.

\ifclearsectionlook\cleardoublepage\fi 
\subsection{Combined Estimation of Blur and Noise}
\label{sec:combinedEstimationResults}
Because previous sections showed that CNN blur and noise estimators performed among the best ones on isolated blur/noise cases, we now use these estimators on combined blur and noise corruption experiments.
Fig.~\ref{fig:combinedEstimationResults} shows the results for combined defocus blur and DCSN (``Defocus + DCSN''), and Photon Shot Noise with simultaneous linear motion blur (``Photon + Lin.~Motion''), both on Udacity.

\subsubsection{Defocus + DCSN} 
According to the physics behind the image formation process in Fig.~\ref{fig:modelBlurNoise}, an image is corrupted by defocus first and DCSN afterwards. 
Hence, high-frequency image content is filtered and fully represented by the DCSN. 
In theory, the larger the blur the easier the noise estimation.
This is what we observe in the first plot of Fig.~\ref{fig:combinedEstimationResults}.
Although there is a small estimation error for zero defocus, $\hat{\sigma}$ becomes most accurate for $d \geq 3 \,\si{\px}$ and remains unchanged. 
Hence, defocus is favorable for DCSN estimation. 
We expect the same effect for other combinations of defocus/motion blur and DCSN/readout noise.

On the other hand, DCSN negatively affects defocus estimation because advantageous information for detecting blur (the absence of high frequencies) gets corrupted by noise.
We notice two effects from the results on the table of Fig.~\ref{fig:combinedEstimationResults}:
All defocus estimations worsen with increasing noise levels, 
and this impact becomes more severe for larger kernel sizes. 
While estimations for the smallest and largest kernels ($d \in \{3, 21\} \,\si{\px}$) can be considered as still good for $\sigma = \SI{10}{\DN}$, the same noise level otherwise leads to poor blur estimations.
This outcome was investigated in the context of motion deblurring \cite{tai2012motion}, where it was found that, as $\sigma$ grows, blur estimations approach the Dirac delta function in a large variety of approaches. 
We observe the same behavior for the CNN estimations, hence the increasing relative error towards larger kernels.
Generally, defocus estimations are not robust in presence of subsequent noise.
Since sensor noise can be detected accurately in case of defocus, a small $\hat{\sigma}$ should be assured before trusting blur estimations. 

\subsubsection{Photon + Lin.~Motion}
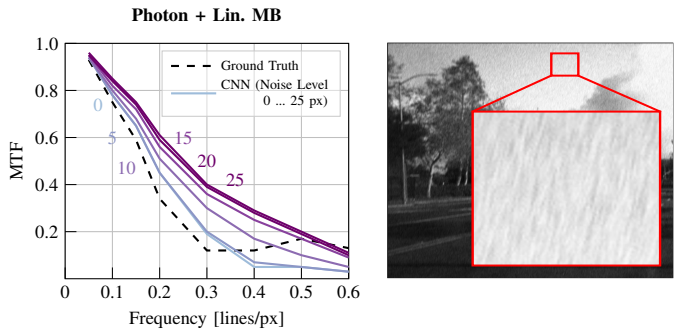
\begin{figure}[t]
    \noindent\begin{minipage}{.28\textwidth}
        \begingroup
            \pgfplotsset{every axis/.style={scale=0.55}}
                \begin{tikzpicture}
    \scriptsize
    \begin{axis}[%
        name=axis1,
        legend cell align={left},
        legend style={
          nodes={scale=0.75, transform shape},
          fill opacity=0.5,
          draw opacity=1,
          text opacity=1,
          at={(0.34,0.97)},
          anchor=north west,
          draw=white!80!black,
          every axis plot/.append style={thick},
          cells={align=left}
        },
        tick pos=left,
        xmajorgrids,
        xmin=0, xmax=0.6,
        xtick style={color=black},
        xtick={0, 0.1, 0.2, 0.3, 0.4, 0.5, 0.6},
        xticklabels={0, 0.1, 0.2, 0.3, 0.4, 0.5, 0.6},
        xlabel={Frequency [lines/px]},
        ymajorgrids,
        ymin=0, ymax=1.0,
        ytick style={color=black},
        ytick={0.2, 0.4, 0.6, 0.8, 1.0},
        yticklabels={0.2, 0.4, 0.6, 0.8, 1.0},
        title={\textbf{Photon + Lin. MB}},
        ylabel={MTF},
        ylabel near ticks,
        colormap/BuPu-7,
        cycle list={[indices of colormap={1,2,3,4,5,6,7 of BuPu-7}]}
    ]
        \addplot[black, dashed] plot coordinates {
        (0.05,0.93) (0.1,0.75) (0.15,0.59) (0.2,0.34) (0.3,0.12) (0.4,0.12) (0.5,0.17) (0.6,0.13)};
        \addlegendentry{Ground Truth};
        \addplot plot coordinates {
        (0.05,0.94) (0.1,0.78) (0.15,0.65) (0.2,0.45) (0.3,0.19) (0.4,0.05) (0.5,0.05) (0.6,0.03)}%
        node at (0.07, 0.74) {0};
        \addlegendentry{CNN (Noise Level \\ \hspace{2.7em} 0 ... 25 px)};
        \addplot plot coordinates {
        (0.05,0.94) (0.1,0.78) (0.15,0.65) (0.2,0.45) (0.3,0.20) (0.4,0.07) (0.5,0.05) (0.6,0.03)}%
        node at (0.1, 0.6) {5};
        \addplot plot coordinates {
        (0.05,0.94) (0.1,0.80) (0.15,0.68) (0.2,0.51) (0.3,0.30) (0.4,0.17) (0.5,0.10) (0.6,0.05)}%
        node at (0.13, 0.47) {10};
        \addplot plot coordinates {
        (0.05,0.95) (0.1,0.82) (0.15,0.72) (0.2,0.56) (0.3,0.36) (0.4,0.25) (0.5,0.17) (0.6,0.09)}%
        node at (0.25, 0.6) {15};
        \addplot plot coordinates {
        (0.05,0.95) (0.1,0.84) (0.15,0.74) (0.2,0.59) (0.3,0.39) (0.4,0.28) (0.5,0.19) (0.6,0.10)}%
        node at (0.3, 0.50) {20};
        \addplot plot coordinates {
        (0.05,0.96) (0.1,0.85) (0.15,0.75) (0.2,0.61) (0.3,0.40) (0.4,0.29) (0.5,0.20) (0.6,0.11)}%
       node at (0.36, 0.42) {25};
       
    \end{axis}
\end{tikzpicture}
        \endgroup
    \end{minipage}%
    \begin{minipage}{.235\textwidth}
        \vspace{-0.2cm}
        \includegraphics[width=0.9\textwidth, trim= 16 16 16 16, clip, height=31.6mm]{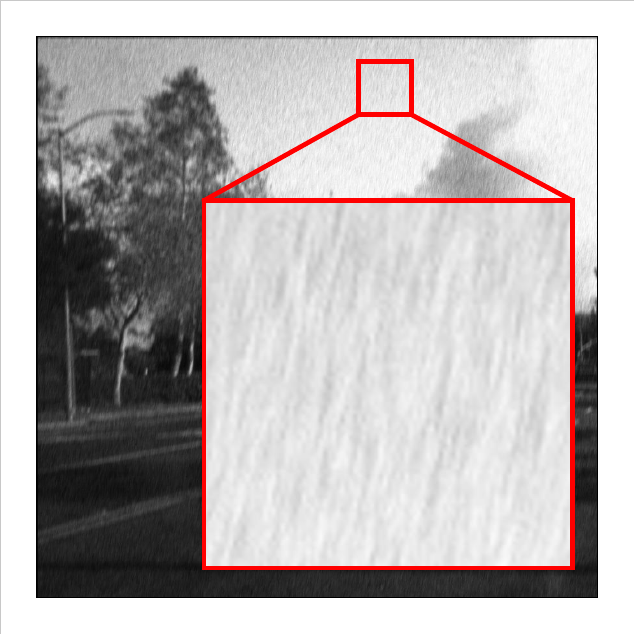}
    \end{minipage}
    \caption{\emph{Linear motion blur estimation in presence of preceding photon shot noise.} Left: Increasing noise levels $\sigma$ increase the MTF estimation and thus decrease the estimated blur level $d$. Right: Corresponding exemplary image with $(d,\sigma) = (\SI{11}{\px}, \SI{25}{\DN})$ showing structured noise induced by subsequent motion blur.}
    \label{fig:blurEstimationInPresenceOfNoiseExample}
    \vspace{-1.5ex}
\end{figure}

In this case noise is added before the blur (due to the physics behind the image formation model in Fig.~\ref{fig:modelBlurNoise}).
Therefore, we expect the opposite behavior, i.e., a poor noise estimation (the blur kernel acts as a classical noise filter) and a good blur estimation.
However, only the noise estimation meets the expectations (see the second plot and table in Fig.~\ref{fig:combinedEstimationResults}).
A motion blur of size $d = \SI{3}{\px}$ already majorly disturbs noise estimation (note that noise is not removed from the image but spread among neighboring pixels).
On the other hand, the motion blur leads to structured directional noise (i.e., false image details), which in turn reduces the estimated blur level by increasing $\widehat{\text{MTF}}$ (Fig.~ \ref{fig:blurEstimationInPresenceOfNoiseExample}). 
This effect intensifies with increasing noise level.
Depending on whether blur is overestimated (e.g., for $d=\SI{3}{\px}$) or underestimated (e.g., for $d=\SI{11}{\px}$) when $\sigma=\si{0}$, the AMAE score decreases or increases for higher noise levels, respectively.

\begin{figure}[t]
    \begin{minipage}{.26\textwidth}
        \centering
        \begin{adjustbox}{max width=1.0\linewidth}
        \setlength{\tabcolsep}{4pt}
        \begin{tabular}{llccccc}
            \toprule
            &&\multicolumn{5}{c}{Photon + Defocus}\\
             \cmidrule(r){3-7}
             Size [$\si{\px}$] && 3 & 7 & 11 & 15 & 21%
             \\
            Kernel && \includegraphics[width=0.085\textwidth]{images/blurKernels/defocus/3.png}%
            & \includegraphics[width=0.085\textwidth]{images/blurKernels/defocus/7.png}%
            & \includegraphics[width=0.085\textwidth]{images/blurKernels/defocus/11.png}%
            & \includegraphics[width=0.085\textwidth]{images/blurKernels/defocus/15.png}%
            & \includegraphics[width=0.085\textwidth]{images/blurKernels/defocus/21.png}%
            \\
            \midrule
            Noise Level
                &0%
                        & 2.7 & 0.9 & 0.6 & 0.3 & 1.4\\
                &5%
                        & 0.6 & 1.1 & 0.8 & 0.4 & 1.3\\
                &10%
                        & 0.5 & 1.1 & 0.5 & 0.4 & 1.3\\
                &15%
                        & 1.2 & 2.5 & 0.6 & 0.4 & 1.3\\
                &20%
                        & 5.4 & 1.7 & 0.5 & 0.3 & 1.4\\
                &25%
                        & 5.6 & 1.8 & 0.5 & 0.3 & 1.4\\
            \bottomrule
        \end{tabular}
        \end{adjustbox}
    \end{minipage}%
    \begin{minipage}{.24\textwidth}
        \hspace{0.255cm}
        \includegraphics[width=0.872\textwidth, trim= 16 16 16 16, clip, height=33.8mm]{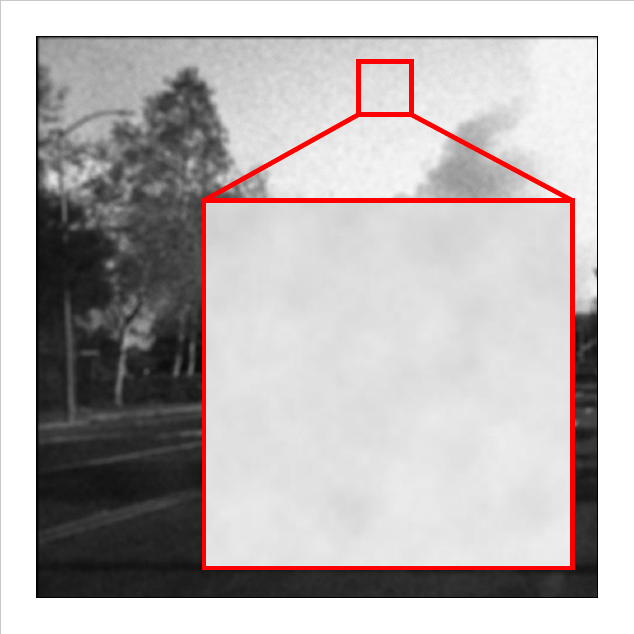}
    \end{minipage}
    \caption{\emph{Defocus blur estimation in presence of preceding photon shot noise.} Left: The minor effect of noise on defocus blur estimation becomes negligible for kernel sizes $d \geq \SI{7}{\px}$. Right: Example with $(d,\sigma) = (\SI{11}{\px}, \SI{25}{\DN})$.}
    \vspace{-2ex}
    \label{fig:blurEstimationPhotonNoise+Defocus}
\end{figure}

We do not observe the same behavior when we replace motion blur with defocus blur (``Photon + Defocus'', Fig.~\ref{fig:blurEstimationPhotonNoise+Defocus}), as defocus blur distributes the noise evenly to the neighboring pixels. The noise still influences the blur estimation of the defocus kernel $d = \SI{3}{\px}$, however, the effect becomes negligible for larger defocus kernels ($d \geq \SI{7}{\px}$).

We build upon this finding and propose a simple approach to suppress high noise in order to re-enable the detection of preceding blur.
Specifically, we apply an additional defocus filter to estimate preceding small or medium blur for high sensor noise levels $\sigma \geq \SI{10}{\DN}$ (details in the supplementary material).
We implemented this improved blur estimation for experiments in Sec.~\ref{sec:sampleApplication}.

In summary, we conclude that even a small amount of blur boosts the detection of subsequent noise while suppressing preceding noise sources. 
So, in the presence of blur, photon noise is difficult to estimate and therefore should be avoided.
Regarding blur estimation, preceding photon noise can corrupt the result in case of motion blur.
Subsequent DCSN with $\sigma \geq \SI{10}{\DN}$ already prevents blur estimation, however, it can be re-enabled by applying an additional defocus filter.
Hence, if one can eliminate photon noise, we suggest estimating noise before judging a blur estimation result.
As in the noise evaluation of Sec.~\ref{sec:noiseEstimationResults}, sensor noise (DCSN and readout noise) is more favorable than photon shot noise for condition monitoring.

\ifclearsectionlook\cleardoublepage\fi \section{Maximizing Object Detection by Trading Off Blur and Noise} 
\label{sec:sampleApplication}%
\begin{figure*}[t]
        \begingroup
            \pgfplotsset{every axis/.style={scale=0.55}}
            \pgfplotsset{cycle list/Set1, cycle multiindex* list={
                    mark list*\nextlist
                    Set1\nextlist
                }
            }
            \input{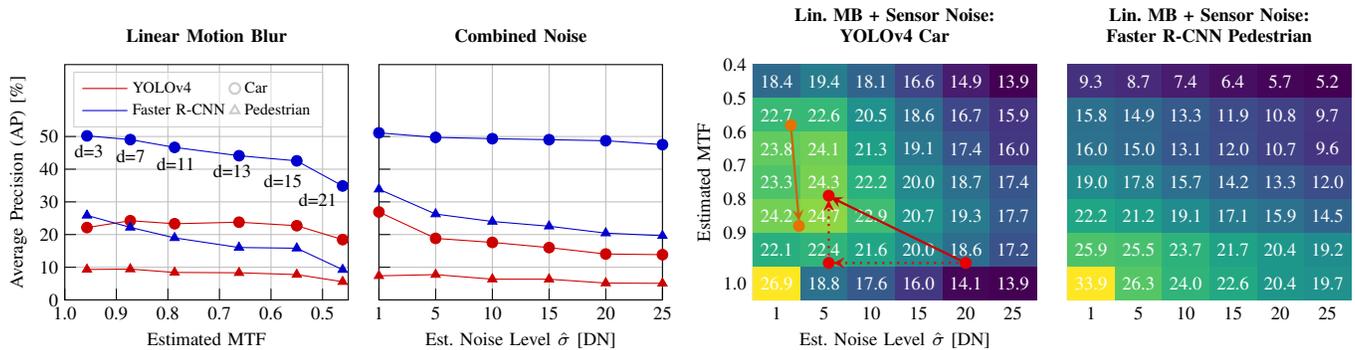}
        \endgroup
    \vspace{-2ex}
    \caption{\emph{Influence of blur and noise on object detection performance.}
    Exemplary object detection performances depending on isolated (Left) and combined (Right) occurring and blur and noise. All input-output profiles depend on the actual estimated corruption levels. Performances are measured in terms of average precision (AP). 
    Noise levels and MTFs are estimated by the respective CNN methods and the MTFs depict means for horizontal and vertical measurements at frequency $f=0.1$. 
    The red and orange arrows demonstrate two examples of exposure time $t_\text{exp}$ / ISO-gain trade-off paths (see text in Sec.~\ref{sec:sampleApplication}).
    }
    \label{fig:applicationMotionBlurPerformanceCurveResults}
\end{figure*}
\begin{figure*}[t]
\centering
    \begin{tikzpicture}
        \node (img) at (0,0) %
        {\includegraphics[width=0.98\linewidth, trim=18 22 18 22, clip]{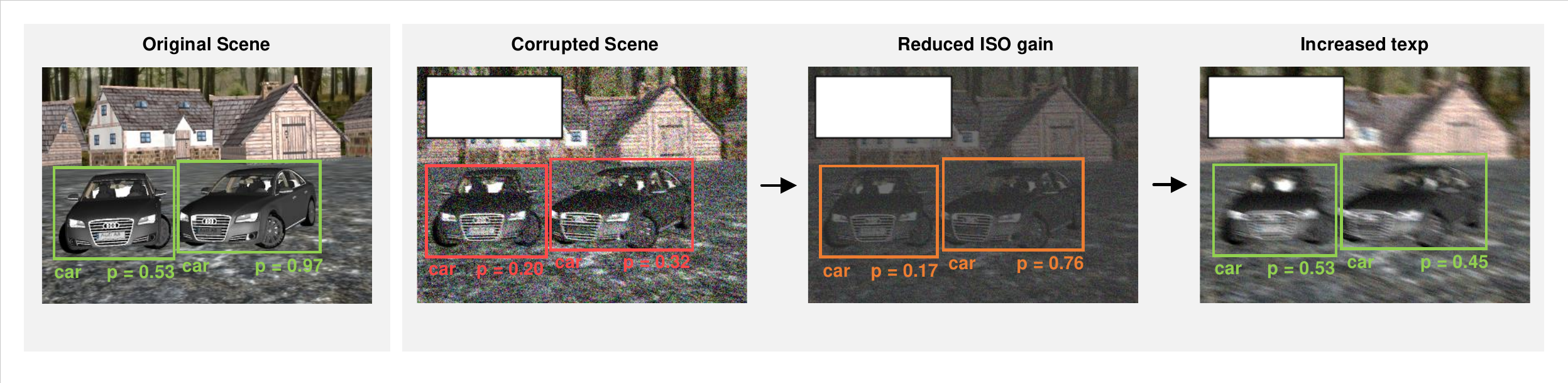}};
        \notsotiny
        \node at (-3.42, 1.10) {$~t_{\text{exp}} ~= \SI{4}{\milli\second}$};
        \node at (-3.42, 0.84) {$\text{ISO} \,= 400$};
        \node at (1.14, 1.10) {$~t_{\text{exp}} ~= \SI{4}{\milli\second}$};
        \node at (1.14, 0.84) {$\text{ISO} \,= 100$};
        \node at (5.74, 1.10) {$~t_{\text{exp}} ~= \SI{16}{\milli\second}$};
        \node at (5.74, 0.84) {$\text{ISO} \hspace{0.07cm}= 100 ~~\,$};
        \small
        \node at (-7.77, -1.67) {$\hat{\sigma} = \SI{1.0}{\DN}$};
        \node at (-5.76, -1.62) {$\widehat{\text{MTF}} = 0.99$};
        \node at (-3.38, -1.67) {$\hat{\sigma} = \SI{20.0}{\DN}$};
        \node at (-1.37, -1.62) {$\widehat{\text{MTF}} = 0.96$};
        \node at (1.24, -1.67) {$\hat{\sigma} = \SI{5.0}{\DN}$};
        \node at (3.25, -1.62) {$\widehat{\text{MTF}} = 0.95$};
        \node at (5.80, -1.67) {$\hat{\sigma} = \SI{5.0}{\DN}$};
        \node at (7.81, -1.62) {$\widehat{\text{MTF}} = 0.71$};
    \end{tikzpicture}%
    \caption{\emph{Maximizing object detection by trading off blur and noise.}
    Application of the proposed framework to detect cars using YOLOv4 on \emph{Sim} data suffering from linear motion blur and sensor noise. 
    The scene (Left) is first imaged with an ISO gain of 400 (leading to sensor noise of $\sigma \approx \SI{20}{\DN}$) and an exposure time of \SI{4}{\milli\second}. 
    As a result, the car recognition performance of YOLOv4 decreases drastically (Center-Left). 
    Applying the optimal $\alpha^\star \approx 0.25$ (according to the performance profile from Fig.~\ref{fig:applicationMotionBlurPerformanceCurveResults}) improves car detection (Center-Right). 
    Finally, we divide the exposure time by $\alpha^\star$ to compensate for the missing light, which improves overall detection slightly (Right). 
    Hence, noise is reduced from $\sigma \approx \SI{20}{\DN}$ to $\SI{5}{\DN}$ and blur is increased from $d \approx \SI{3}{\px}$ to $\SI{12}{\px}$ while detection rate increases from $p\approx 0.25$ to $p\approx 0.5$.
    }
    \label{fig:applicationMotionBlurSimulationResults}
\end{figure*}
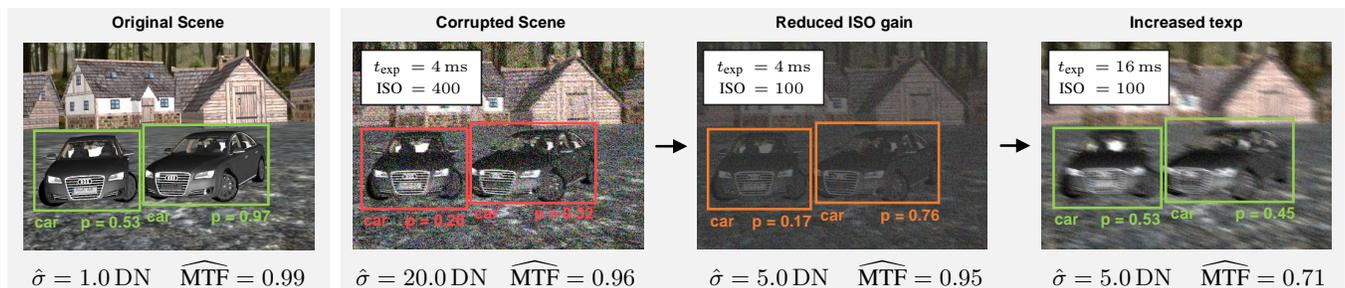

\begin{figure}[t]
\centering
    \includegraphics[width=0.75\linewidth, trim=20 18 18 18, clip]{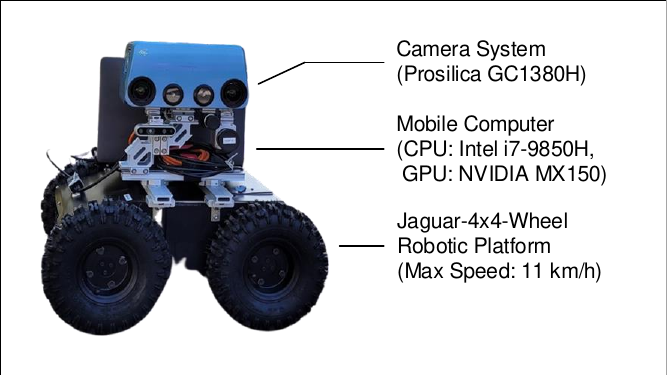}
    \caption{\emph{Deploying our framework on a real camera system} that is attached on an autonomous mobile robotic platform.}
    \label{fig:roboticPlatform}
    \vspace{-2ex}
\end{figure}

In this section we demonstrate the application of the proposed framework in a simulated and a real-world scenario (Secs.~\ref{sec:example1} and \ref{sec:example2}, respectively).

Let us first calculate exemplary IOPCs according to Sec.~\ref{sec:imgQualityAssessmentPerformanceProfiles} with focus on:
($i$) object classes \emph{car} and \emph{pedestrian}, 
($ii$) sensor noise only (DCSN + read noise), so that filtered photon noise does not lower the noise level estimation,
and ($iii$) settings from Sec.~\ref{sec:experim:datasets} for linear motion blur and sensor noise generation.
Figure \ref{fig:applicationMotionBlurPerformanceCurveResults} shows the resulting IOPCs for isolated (left) and combined (right) blur and noise occurrences. 
It can be seen that the relation from the blur or noise corruptions to the detection performances might be non-trivial and non-linear (e.g., YOLOv4 car detection in presence of Lin.~MB) 
since it is difficult to tell what ML methods learn.

While the proposed IOPC approach is data-driven, it is possible to obtain a model of the object detection AP performance in terms of exposure time ($t_\text{exp}$) and ISO gain ($A_\text{ISO}$) camera parameters.
To this end, we fitted a multivariate function to the IOPCs in Fig.~\ref{fig:applicationMotionBlurPerformanceCurveResults}, with appropriate basis functions to avoid overfitting while accounting for the IOPSs' non-linearity and non-monotonicity:
\begin{equation}
\label{eq:detectionPerformanceByExposureTimeAndISOGain}
    \widehat{\text{AP}} = 
    c_1 xy + c_2 x + c_3 y + c_4 \sqrt{y} + c_5
    + \frac{c_6}{x y} + \frac{c_7}{x} + \frac{c_8}{y},  
\end{equation}
with the fitted detection score $\widehat{\text{AP}}$, $x \propto A_\text{ISO}$, $y \propto t_\text{exp}$, and regression coefficients $c_1, \dots, c_8$.
The details of this fitting process are provided in the supplementary material.

\subsection{Example 1}
\label{sec:example1}%
We first demonstrate this framework using the \emph{Sim} environment on a concrete example of YOLOv4 car detection with corrupted data by means of Lin.~MB and sensor noise (Fig.~\ref{fig:applicationMotionBlurSimulationResults}).
The left image in Fig.~\ref{fig:applicationMotionBlurSimulationResults} depicts the scene in uncorrupted conditions (without noise or blur), for reference. 
Here the first car is detected fairly ($p = 0.53$) and the second one much better ($p = 0.97$).
While the CNN noise estimator detects a small noise level of $\hat{\sigma} = \SI{1}{\DN}$ by mistake, the MTF estimation is nearly ideal ($\widehat{\text{MTF}} = 0.99$).
Next, we included a realistic trajectory for the simulated camera to create a linear motion with a speed of $v \approx \SI{760}{\px/\second}$. %
This causes blur (an exposure time of $t_{\text{exp}} = \SI{4}{\ms}$ induces a motion blur of $d_{\text{old}} \approx \SI{3}{\px}$), and we also apply sensor noise of $\sigma = \SI{20}{\DN}$.
In this situation (second image in Fig.~\ref{fig:applicationMotionBlurSimulationResults}) blur and noise are  estimated within the expected error ranges ($\hat{d}_{\text{old}} \approx \SI{3}{\px}$), but the cars are detected worse ($p\approx 0.25$). %

In the next step, we determine $\alpha^\star$:
knowing the relation between motion blur sizes and estimated MTFs (first plot in Fig.~\ref{fig:applicationMotionBlurPerformanceCurveResults}) and the estimated noise level, we target an $\widehat{\text{MTF}} \in [0.7, 0.8]$, which corresponds to approximately $\hat{d}\in [11, 12] \,\si{\px}$ (cf. first heat plot in Fig.~\ref{fig:applicationMotionBlurPerformanceCurveResults}). 
We chose $d_{\text{target}} \approx \SI{12}{\px}$, hence, $\alpha^\star = \hat{d}_{\text{old}} / d_{\text{target}} \approx 3 / 12 = 0.25$.
We then reduce the ISO gain by the factor $\alpha^\star$ and show an intermediate image without increasing $t_\text{exp}$.
One car is now detected more confidently while blur and noise are still estimated within the expected error ranges.
As we did not investigate the influence of image intensity on object detection performance, next we increase $t_\text{exp}$ by the factor $\alpha^\star$ to restore the original intensity level, producing the last image of Fig.~\ref{fig:applicationMotionBlurSimulationResults}. 
In this last step the total detection score slightly increases despite the likewise motion blur amplification ($d \approx \hat{d} = \SI{12}{\px}$).
The steps taken are marked with red arrows on the heat plot in Fig.~\ref{fig:applicationMotionBlurPerformanceCurveResults}.

\subsection{Example 2}
\label{sec:example2}%
\begin{figure}[t]
\centering
    \begingroup
        \pgfplotsset{every axis/.style={scale=0.59}}
        \begin{tikzpicture}
    \small
    \begin{axis}[%
        name=axis1,
        xshift=+4mm,
        legend cell align={left},
        legend columns=-1,
        legend style={
          nodes={scale=0.85, transform shape},
          fill opacity=1.0,
          draw opacity=1,
          text opacity=1,
          at={(0.01,0.97)},
          anchor=north west,
          draw=white!80!black,
          /tikz/every even column/.append style={column sep=0.5cm}
        },
        legend entries={Camera Default, Proposed},
        tick pos=left,
        xmajorgrids,
        xmin=0, xmax=5000,
        xtick style={color=black},
        xtick={0, 1000, 2000, 3000,  4000, 5000},
        xlabel={Time $t$ [ms]},
        ymajorgrids,
        ymin=0, ymax=25,
        width=12cm,
        height=6cm,
        ytick style={color=black},
        ytick={0, 5, 10, 15, 20, 25},
        ylabel={\color{blue!80!black}{Exposure Time} \\ \color{blue!80!black}{$t_\text{exp}$ [ms]}},
        ylabel near ticks,
        ylabel style={align=center},
        title style={align=center},
        title={\footnotesize \textbf{Camera Parameter Adjustment Response Time}}
    ]
        \addlegendimage{gray, thick, dashed}
        \addlegendimage{gray, thick}
        
        \addplot [blue!80!black, thick, dashed] coordinates {
        (0, 1)
        (100, 2.65)
        (200, 3.817)
        (300, 4.795)
        (400, 5.74)
        (500, 6.486)
        (600, 7.329)
        (700, 9.20799999999999)
        (800, 11.568)
        (900, 13.301)
        (1000, 14.187)
        (1100, 14.632)
        (1200, 14.632)
        (1300, 10.974)
        (1400, 11.5669999999999)
        (1500, 11.814)
        (1600, 11.814)
        (1700, 11.814)
        (1800, 11.814)
        (1900, 12.067)
        (2000, 13.419)
        (2100, 14.696)
        (2200, 15.491)
        (2300, 15.491)
        (2400, 16.143)
        (2500, 16.143)
        (2600, 16.649)
        (2700, 17.005)
        (2800, 17.005)
        (2900, 17.005)
        (3000, 17.005)
        (3100, 17.005)
        (3200, 17.005)
        (3300, 17.005)
        (3400, 17.369)
        (3500, 13.027)
        (3600, 14.073)
        (3700, 15.203)
        (3800, 15.843)
        (3900, 16.182)
        (4000, 16.528)
        (4100, 16.882)
        (4200, 16.882)
        (4300, 16.882)
        (4400, 16.882)
        (4500, 16.882)
        (4600, 17.243)
        (4700, 12.933)
        (4800, 13.027)
        (4900, 14.151)
        (5000, 14.685)
        };
        
        \addplot [blue!80!black, thick] coordinates {
        (0, 8)
        (100, 8)
        (200, 8)
        (300, 8)
        (400, 8)
        (500, 8)
        (600, 8)
        (700, 8)
        (800, 8)
        (900, 8)
        (1000, 8)
        (1100, 8)
        (1200, 8)
        (1300, 8)
        (1400, 8)
        (1500, 8)
        (1600, 8)
        (1700, 8)
        (1800, 8)
        (1900, 8)
        (2000, 8)
        (2100, 8)
        (2200, 8)
        (2300, 8)
        (2400, 8)
        (2500, 8)
        (2600, 8)
        (2700, 8)
        (2800, 8)
        (2900, 8)
        (3000, 8)
        (3100, 3.733)
        (3200, 3.733)
        (3300, 3.733)
        (3400, 3.733)
        (3500, 3.733)
        (3600, 3.733)
        (3700, 3.733)
        (3800, 3.733)
        (3900, 3.733)
        (4000, 3.733)
        (4100, 3.733)
        (4200, 3.733)
        (4300, 3.733)
        (4400, 3.733)
        (4500, 3.733)
        (4600, 3.733)
        (4700, 3.733)
        (4800, 3.733)
        (4900, 3.733)
        (5000, 3.733)
        };
    \end{axis}
    
    \pgfplotsset{every axis y label/.append style={rotate=180}}
    \begin{axis}[
        xmin=0, xmax=5000,
        ymin=0, ymax=400,
        xshift=0.4cm,%
        hide x axis,
        height=6cm,
        width=12cm,
        axis y line*=right,
        ytick = {100, 200, 300},
        ylabel={\color{red!80!black}{ISO Gain}}
    ]
    
    \addplot [red!80!black, thick, dashed] coordinates {
        (0, 100)
        (100, 100)
        (200, 100)
        (300, 100)
        (400, 100)
        (500, 100)
        (600, 100)
        (700, 100)
        (800, 100)
        (900, 100)
        (1000, 100)
        (1100, 100)
        (1200, 100)
        (1300, 100)
        (1400, 100)
        (1500, 100)
        (1600, 100)
        (1700, 100)
        (1800, 100)
        (1900, 100)
        (2000, 100)
        (2100, 100)
        (2200, 100)
        (2300, 100)
        (2400, 100)
        (2500, 100)
        (2600, 100)
        (2700, 100)
        (2800, 100)
        (2900, 100)
        (3000, 100)
        (3100, 100)
        (3200, 100)
        (3300, 100)
        (3400, 100)
        (3500, 100)
        (3600, 100)
        (3700, 100)
        (3800, 100)
        (3900, 100)
        (4000, 100)
        (4100, 100)
        (4200, 100)
        (4300, 100)
        (4400, 100)
        (4500, 100)
        (4600, 100)
        (4700, 100)
        (4800, 100)
        (4900, 100)
        (5000, 100)
        };
        
        \addplot [red!80!black, thick] coordinates {
        (0, 100)
        (100, 100)
        (200, 100)
        (300, 100)
        (400, 100)
        (500, 100)
        (600, 100)
        (700, 100)
        (800, 100)
        (900, 100)
        (1000, 100)
        (1100, 100)
        (1200, 100)
        (1300, 100)
        (1400, 100)
        (1500, 100)
        (1600, 100)
        (1700, 100)
        (1800, 100)
        (1900, 100)
        (2000, 100)
        (2100, 100)
        (2200, 100)
        (2300, 100)
        (2400, 100)
        (2500, 100)
        (2600, 100)
        (2700, 100)
        (2800, 100)
        (2900, 100)
        (3000, 100)
        (3100, 224)
        (3200, 224)
        (3300, 224)
        (3400, 224)
        (3500, 224)
        (3600, 224)
        (3700, 224)
        (3800, 224)
        (3900, 224)
        (4000, 224)
        (4100, 224)
        (4200, 224)
        (4300, 224)
        (4400, 224)
        (4500, 224)
        (4600, 224)
        (4700, 224)
        (4800, 224)
        (4900, 224)
        (5000, 224)
        };
    
    \end{axis}
\end{tikzpicture}
    \endgroup%
    \begin{tikzpicture}
        \node (img) at (0,0) %
        {\includegraphics[width=0.96\linewidth, trim=16 16 16 16, clip]{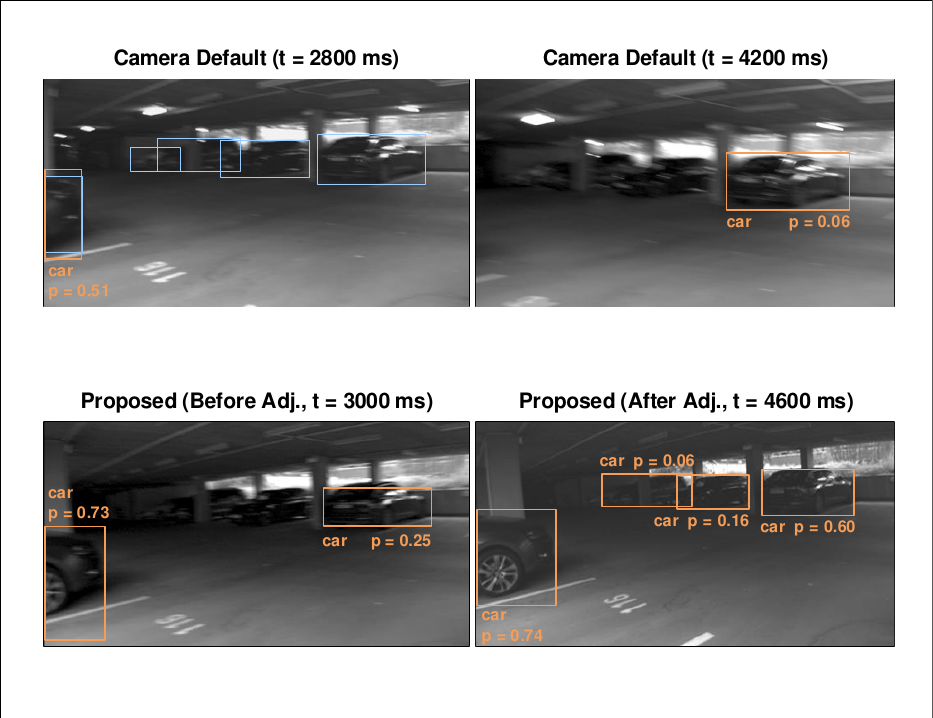}};
        \small
        \node at (-3.24, 0.09) {$\hat{\sigma} = \SI{0.3}{\DN}$};
        \node at (-1.09, 0.15) {$\widehat{\text{MTF}} = 0.71$};
        \node at (1.10, 0.09) {$\hat{\sigma} = \SI{0.3}{\DN}$};
        \node at (3.20, 0.15) {$\widehat{\text{MTF}} = 0.61$};
        \node at (-3.24, -3.28) {$\hat{\sigma} = \SI{0.2}{\DN}$};
        \node at (-1.09, -3.23) {$\widehat{\text{MTF}} = 0.57$};
        \node at (1.10, -3.28) {$\hat{\sigma} = \SI{0.4}{\DN}$};
        \node at (3.20, -3.23) {$\widehat{\text{MTF}} = 0.90$};
    \end{tikzpicture}%
    \caption{\emph{Comparison of built-in camera $t_{\text{exp}}$ / gain control vs.~our framework.} 
    Diagram: The built-in controller constantly optimized image intensity by adjusting $t_{\text{exp}}$ only. 
    In contrast, our framework targets optimal object detection performance and adjusted $t_{\text{exp}}$ / gain once. 
    Images: Examples from the experiments (adapted brightness and contrast for better visualization). Blue boxes indicate ground truth objects, and orange boxes actual detections. 
    The overall AP scores were: 
    26.08\% for the built-in camera control, 
    47.54\% for the manually chosen fixed parameters (before automatic adjustment), 
    and 60.56\% for our framework (after automatic adjustment).}
    \vspace{-2ex}
    \label{fig:applicationMotionBlurRealnResults}
\end{figure}
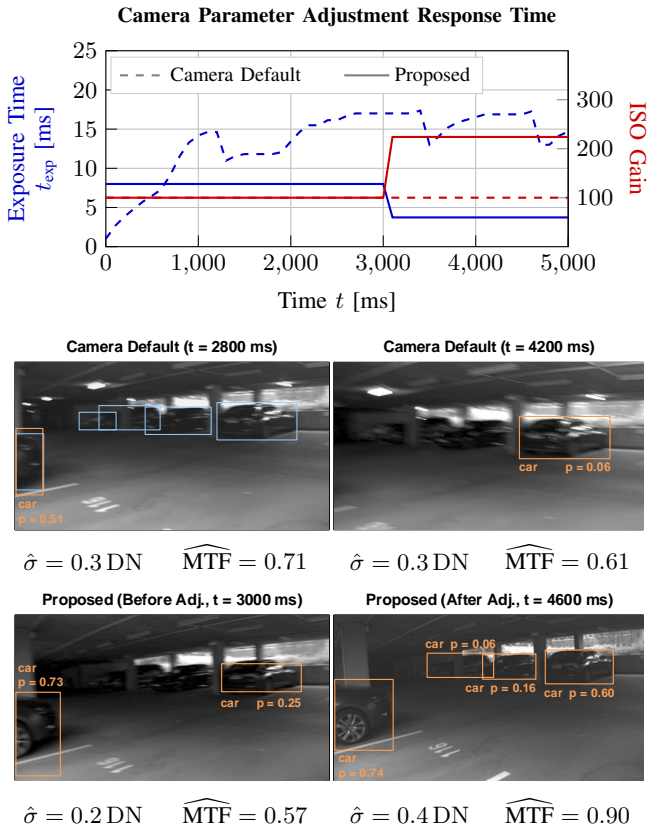

For a real-world example, we deployed our framework on a real camera system using an Allied Vision Prosilica GC1380H camera \cite{prosilicaGC1380HAPI} attached on a Jaguar-4x4-wheel mobile robotic platform \cite{drRobot4x4Wheel} and a mobile computer doing the real-time calculations (Fig.~\ref{fig:roboticPlatform}).
With this setup, we demonstrate another example of the non-monotonic YOLOv4--car-detection heat map of Fig.~\ref{fig:applicationMotionBlurPerformanceCurveResults}, marked with an orange arrow. 
We therefore navigated the camera system through a low-illuminated parking lot with fixed initial camera parameters $t_{\text{exp}} = \SI{8}{\milli\second}$, $\text{ISO} = 100$, and default values for the rest. 
To make the YOLOv4--car-detection profile applicable, we target ($i$) a constant linear motion blur induced by a constant speed of the platform and ($ii$) a low noise level with the low ISO gain to neglect the undesired impact of photon shot noise.
Subsequently, the experiment was repeated with the camera's built-in $t_{\text{exp}}$ / ISO gain controller \cite{prosilicaGC1380HAPI} and compared to ours with respect to response time and object detection performance (Fig.~\ref{fig:applicationMotionBlurRealnResults}). 
Both camera parameter controllers were automatically triggered on the first image frame at $t = \SI{0}{\milli\second}$.
Finally, we sampled the video sequences for each configuration to contain $200 \pm 5$ cars, which we manually annotated for YOLOv4.

The built-in camera controller tracks a mean image intensity level of 50\% and prioritizes changing $t_{\text{exp}}$ over ISO gain as long as $t_{\text{exp}} \leq \SI{500}{\milli\second}$. Hence, the built-in controller constantly changed $t_{\text{exp}}$ only, did not account for the motion blur, and resulted in an AP car detection score of 26.08\%.

Our proposed framework took about \SI{3000}{\milli\second} to estimate $(\hat\sigma,\widehat{\text{MTF}}) = (0.1, 0.57)$ (longer than in Sec.~\ref{sec:experiments} due to the weaker mobile hardware, but still interactive / real-time). 
With initially fixed camera parameters (i.e., while $t < \SI{3000}{\milli\second}$), YOLOv4 reached an AP score of 47.54\%. The system then decided to decrease the motion blur at the expense of slightly increasing the noise to move to higher AP detection values (brighter part of the heat map). 
Inspecting the AP curves (1st plot in Fig.~\ref{fig:applicationMotionBlurPerformanceCurveResults}), $\text{MTF}=0.57$ corresponds to $d=\SI{15}{\px}$, and the system targeted $\widehat{\text{MTF}} \in [0.8, 0.9]$ (high values of the heat map), which corresponds to a smaller motion blur of $\hat{d} \approx \SI{7}{\px}$. 
Two steps were taken: first, the system decreased the exposure time by a factor $\alpha = 15/7 \approx 2.14$ to achieve the desired MTF improvement. Intel
Then, it increased the ISO (and increased noise) by the same factor $\alpha \approx 2.14$ to restore the intensity level for the detector. 
The final operating point was $(\hat\sigma,\widehat{\text{MTF}}) \approx (0.4, 0.9)$, which has a higher AP value (60.56\%) than the initial point.

\ifclearsectionlook\cleardoublepage\fi \section{Conclusion}
\label{sec:conclusion}

We have proposed a framework for real-time camera conditioning, bringing together the tasks of inferring the state of the system and acting on the camera's operating point to achieve optimal application performance. 
Our framework has a modular design, hence it is flexible and interpretable, allowing for multiple choices of its submodules, such as the image quality estimators.
To this end, we have carried out a comprehensive experimental study close to the physics of the sensor %
and have incorporated six state-of-the-art image quality estimators, two advanced object detectors and two standard datasets plus one self-created. 
We have considered a more extensive and realistic image formation pipeline than preceding works by including motion and defocus blur as well as simultaneous occurring corruptions that influence each other.
All these elements have been put together in a coherent manner to justify our design choices and provide insights and practical recommendations with regard to camera monitoring applications (summarized at the end of each experimental subsection).

Regarding the framework, the main idea is that aiming at improving image quality blindly, without taking into account the subsequent high-level application, may not always be the best. 
If the end goal is better high-level application performance (say, car detection), then it is sensible to trade off image quality for whole system performance by adjusting the camera parameters.
We have demonstrated this on how image blur and noise (image quality) affect an object detection application and have implemented it on a real-world ground robot;
the specific control strategy of the camera parameters (exposure time and ISO gain) depends on the experimental input-output performance curves of the object detector (which is in general non-linear and non-monotonic).
However, our framework is generic: %
it is not limited to the proposed control strategy (one could control a motor to adjust focus), %
it can be applied to other optical sensor systems (as infrared or event cameras), %
other scenarios, %
and it can consider other ``features'', conceivably application-specific, besides blur and noise. 
These have been selected because they are among the most generic and influential effects in image processing.
The framework can also be easily extended to optimize multiple vision applications running simultaneously (e.g., by maximizing their weighted average performances).

Lastly, we have focused on a subset of corruptions originating in the camera itself.
A possible extension would be to model additional camera conditions and retrain the estimators accordingly (e.g., more sophisticated corruptions such as defocus due to heat-induced material stress or conditions originating outside the camera, like low/high scene illumination), and compensate for them by triggering actuators (e.g., cooling or headlights).
This could also require the acquisition and exploitation of additional data, such as the camera's and environment's configuration (focal length, aperture size, exposure time, temperature, positioning, illumination), leading to the research and development of more advanced Sensor AI approaches~\cite{sensorAIWhitepaper}.

\bibliographystyle{IEEEtran}
\ifarxiv
\else 

\fi 

\cleardoublepage \section*{Supplementary Material}
The following supplementary material complements the main paper.
We first detail the assumed image formation process (Sec.~\ref{sec:fundamentals}) and subsequently propose a simple approach to improve blur estimations in the presence of high noise (Sec.~\ref{sec:improvedBlurEstimationInPresenceOfNoise}).
Finally, we provide two additional experiments for the sake of completeness: a performance curve analysis of the latest YOLOv7 object detector (Sec.~\ref{sec:yolov7}) and a data-driven model derivation for YOLOv4 object detection performance as a function of exposure time and ISO gain camera parameters (Sec.~\ref{sec:objectDetectionPerformanceDependingOnCameraParameters}).
\section{Image Formation Process} \label{sec:fundamentals}
The image formation process that we consider in a standard camera is depicted in Fig.~\ref{fig:modelBlurNoise}. 
Let us specify the image blur and noise components of this model, and metrics to quantify them.

\subsection{Blur} \label{sec:fundamentalsBlur}
Image blur is the result of processes that reduce image sharpness. 
The most prominent of such processes are ($i$) light refracted by a defocused lens, ($ii$) motion between the sensor and the scene, ($iii$) atmospheric turbulence, and ($iv$) diffraction \cite[p.~325]{bookJayaraman2009}. 
We focus on the former two sources, whose induced blur types are known as defocus and motion blur, respectively. 
Many factors contribute to these processes and make their mathematical description complex. 
For the sake of simplicity they are often modelled as a convolution on the image plane:
\begin{equation} \label{eq:blur}
I^{*}(x,y) = I(x,y) \circledast h(x,y),
\end{equation}
where $I(x,y)$ is the input intensity at pixel $(x,y)$ (before the blur process),
$h(x,y)$ is the blur kernel and $I^{*}(x,y)$ is the blurred image intensity.
The kernel $h(x,y)$ is also called point spread function (PSF)~\cite[p.~328]{bookJayaraman2009}.

The PSF can be used to objectively quantify image blur. 
Its Fourier transform is the optical transfer function (OTF) and it describes how spatial frequencies $f$ (i.e., image details, contrast) are affected by blur:
\begin{equation}
\label{eq:psf}
\PSF(x,y) \quad\stackrel{\mathcal{F}}{\mapsto}\quad \OTF(f) \propto \MTF(f) \smallSpace e^{i \smallSpace \PhTF(f)}.
\end{equation}
Usually only the magnitude of the OTF, known as the modulation transfer function (MTF), is relevant to quantify blur, and so the phase transfer function (PhTF) is omitted.
Let us now describe defocus and motion blur kernels $h(x,y)$.

\subsubsection{Defocus Blur}
We assume a defocus blur kernel $h(x,y)$ that distributes a pixel's intensity evenly over an approximate circular area of neighboring pixels (with radius $r$ and center $(c_x,c_y)$) \cite[p.~325]{bookJayaraman2009}: %
\begin{equation}
    \begin{aligned} \label{eq:defocusBlur}
         h(x,y) = \smallSpace &
         \begin{cases}
                s, & (x-c_x)^2 + (y-c_y)^2 \leq r^2\\
                0, & \text{otherwise},
         \end{cases}
    \end{aligned}
\end{equation}
with the value $s$ determined by the normalization constraint $\iint h(x,y) \smallSpace dx\smallSpace dy = 1$.
This circle refers to the term circle of confusion, whose diameter $d = 2r+1$ can be calculated as
\begin{equation} \label{eq:defocusBlurCoC}
d = A \smallSpace \frac{\focal}{S_1 - \focal} \smallSpace \frac{\left| S_2 - S_1\right|}{S_2},
\end{equation}
expressed in terms of the focused object distance ($S_1$), the out-of-focus object distance ($S_2$), the focal length ($\focal$), the image distance ($\focal_1$) and the aperture diameter ($A$) \cite[p.~216]{defocusBlurCoC}. 
This defocus blur kernel model has shown to be on par with more complex models in image reconstruction \cite{savakis1993accuracy}.
We assume the camera comprises a single, perfect, convex, thin lens satisfying $1/\focal = 1/\focal_1 + 1/S_1$.

\subsubsection{Motion Blur}
\label{sec:fundamentalsBlurMotion}%
Depending on the type of motion, image blur can manifest as translation, rotation, scale changes or a combination of all of them. 
Hence, a closed-form expression for $h(x,y)$ may be complex to obtain. 
Its main influencing factors are the exposure time and the relative angular speed between the imaged objects and the sensor during the exposure (see \cite[p.~326]{bookJayaraman2009} for an exemplary approximation of $h$ in a simple scenario). 
We model $h(x,y)$ to contain a coherent path of pixels with non-zero and potentially inhomogeneous intensities. 
We assume the path in $h(x,y)$ may be non-linear, since factors like an uneven driving ground and unpredictable moving scene objects might lead to complex non-linear movements during the exposure interval.
For simplicity we neglect additional influences like the camera's readout procedure, the influence of the shutter or rapidly changing lightning conditions.

\subsection{Noise} \label{sec:fundamentalsNoise}

Image noise denotes ``any undesired information that contaminates an image'' and often occurs during image acquisition or transmission \cite[p.~348]{bookJayaraman2009}. 
Having the online condition monitoring approach in mind, we tackle the problem of online characterization and mitigation of image acquisition noise. 
We consider time-varying sources because time-invariant noise sources (such as photo response non-uniformity) are often addressed during calibration (before acquisition) and their residuals are assumed to have a minor influence on image quality. 
Generally, noise can be modelled by:
\begin{equation} \label{eq:generalNoise}
\tilde{I}(x,y) = I(x,y) + I(x,y)^{\gamma} \smallSpace u(x,y)\text{,}
\end{equation}
where $I(x,y)$ is the clean intensity (the signal's intensity), $u(x,y)$ is a random, stationary and uncorrelated noise process, 
and $\tilde{I}(x,y)$ is the corrupted intensity. A parameter $\gamma$ controls different noise types.
The amount of noise (or noise level) may be quantified using the standard deviation $\sigma$ of the underlying statistical distribution of $u(x,y)$. %

Let us now detail the most prominent time-varying noise processes, namely photon shot noise, dark shot noise and readout noise (Fig.~\ref{fig:modelBlurNoise}). 
As a theoretical guide, we follow \cite{noiseSimulation}. 

\subsubsection{Photon Shot Noise}
As photons arrive at the sensor, the counting process within the exposure interval undergoes random fluctuations.
This is known as shot noise and follows a Poisson distribution. 
If the number of arriving photons $k$ is large enough (i.e., in non-low illumination conditions), the Poisson distribution may be approximated by a Gaussian distribution using the Central Limit Theorem \cite[p.~225]{poissonToGaussDistr}:
\begin{equation} \label{eq:poissongToGaussDistr}
    \mathcal{P}_{\lambda} (k) = \frac{\lambda^k}{k!} e^{-\lambda} ~\stackrel{k \, \rightarrow \, \infty}{\approx}~
    \frac{1}{\sqrt{2\pi\lambda}} \smallSpace e^{-(k-\lambda)^2 / 2\lambda}\text{,}
\end{equation}
with $\lambda = \sigma^2$ as the expected value and variance of the arrival events.
The higher the number of arriving photons, the higher the number of random fluctuations; 
hence photon shot noise behaves signal-dependent and can be described by \eqref{eq:generalNoise} when setting $\gamma = 1$ and $u(x,y) \sim P_{\lambda}(k)$.

\subsubsection{Dark Current Shot Noise (DCSN)} 
Similar to photon shot noise, dark current (DC) shot noise originates from the random arrival of DC electrons and follows the same distribution \eqref{eq:poissongToGaussDistr}. 
DC emerges from thermally generated electrons at different sensor material regions.
The amount of generated electrons depends, among others, mainly on the pixel area, temperature and exposure time \cite[ch.~7.1.1]{scientificCCD}.
DCSN is signal-independent, hence $\gamma = 0$ in \eqref{eq:generalNoise}.

\subsubsection{Readout Noise}
Readout noise refers to the imperfections due to the sensor's electronic circuitry converting charge into digital values 
and it is attributed to the on-chip amplification and conversion processing units \cite[p.~197]{readNoise}. 
Although readout noise can be reduced to a negligible level in scientific cameras, its impact is still significant for industry-grade sensors that lack of noise reduction \cite[ch.~7.2.9]{scientificCCD}. 
We incorporate sense node reset noise (alias kTC noise) and source-follower noise as the main time-varying components.

Both noise sources can be modelled as a zero-mean Gaussian process, where $\sigma$ mainly depends on the temperature. 
The overall readout noise contribution is a signal-independent addition of both noise processes, hence $\gamma = 0$ in \eqref{eq:generalNoise}.
We keep it at this level of abstraction and refer to \cite{noiseSimulation} for details.

In summary, we consider the blur and noise sources in Fig.~\ref{fig:modelBlurNoise} and have described their physical models mathematically.

\section{Improved Blur Estimation in Presence of High Noise}
\label{sec:improvedBlurEstimationInPresenceOfNoise}
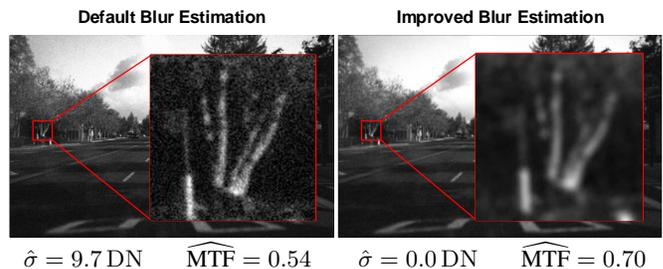
\begin{figure}[t]
    \centering
        \begin{tikzpicture}
        \node (img) at (0,0) %
        {\includegraphics[width=0.99\linewidth, trim=16 16 16 16, clip]{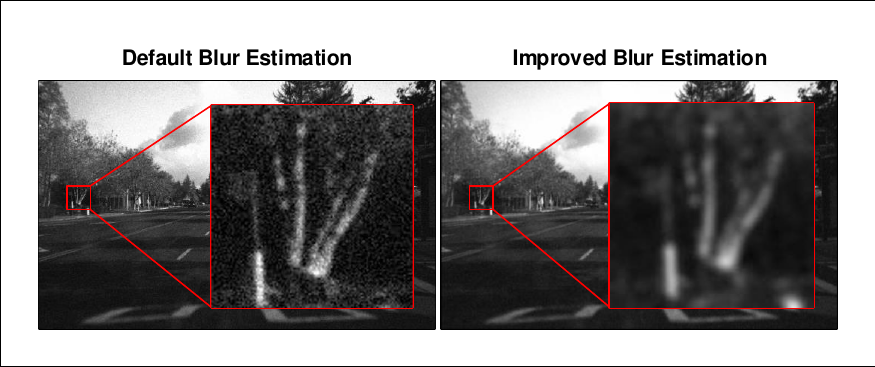}};
        \small
        \node at (-3.35,-1.86) {$\hat{\sigma} = \SI{9.7}{\DN}$};
        \node at (-1.15,-1.81) {$\widehat{\text{MTF}} = 0.54$};
        \node at (1.1,-1.86) {$\hat{\sigma} = \SI{0.0}{\DN}$};
        \node at (3.3,-1.81) {$\widehat{\text{MTF}} = 0.70$};
    \end{tikzpicture}
    \caption{\emph{Proposed improved blur estimation in presence of high noise,} in an exemplary scenario of Lin.~MB + DCSN with $d_1=\SI{11}{\px}$ and $\sigma=\SI{10}{\DN}$. 
    We target estimating the lin.~MB with a ground truth of $\text{MTF}^\text{GT} = 0.75$ (combined image directions at $f=0.1$). 
    Left: Noise distracts the blur estimation ($\widehat{\text{MTF}} = 0.54$). Right: Defocus filtering the noise with $d_2=\SI{7}{\px}$ assists the blur estimation ($\widehat{\text{MTF}} = 0.70$, the influence of defocus was cancelled out during the estimation).}
    \label{fig:improvedBlurEstimationExample}
\end{figure}

The Sec.~\ref{sec:combinedEstimationResults} has pointed out that blur is not accurately estimated in the case of high subsequent noise (e.g., DCSN, with $\sigma \geq \SI{10}{\DN}$).
Here we demonstrate a simple approach to improve the accuracy of such MTF estimates (Fig.~\ref{fig:improvedBlurEstimationExample}).
The approach exploits that preceding photon noise is not expected to significantly influence the MTF estimation of subsequent defocus blur (see Sec.~\ref{sec:combinedEstimationResults}).
Hence, the approach consists of considering the above-mentioned ``high subsequent noise'' as the preceding noise of a new blur stage, estimating the overall MTF and reassigning the credit between the two blur stages.
Specifically, following up on the Defocus + DCSN case in Sec.~\ref{sec:combinedEstimationResults}, the considered pipeline has now three stages: Lin.~MB + DCSN + defocus filtering.
Letting the first blur kernel be $b_1$, we filter noise by an additional kernel $b_2$, estimate the overall blur $\widehat{\text{MTF}}(b_1, b_2) = \widehat{\text{MTF}}(b_1)\, \widehat{\text{MTF}}(b_2)$ and lastly divide the MTF by the known $\text{MTF}^{\text{GT}}(b_2)$ according to the Fourier convolution theorem \cite[p.~242]{jahne2000computer}. 
To this end, we assume $\text{MTF}^{\text{GT}}(b_2) \approx \widehat{\text{MTF}}(b_2)$ and determine the estimation error of $\widehat{\text{MTF}}(b_1)$ with respect to $\text{MTF}^{\text{GT}}(b_1)$.

Due to the combinatorial complexity of the experimental configuration, we focus on the following one grounded in the results from Sec.~\ref{sec:combinedEstimationResults}:
We employ only the CNN method for MTF estimation on Udacity data, 
and in preparation for Sec.~\ref{sec:sampleApplication}, we consider the case of Lin.~MB + DCSN (representative for sensor noise, to keep it clear and concise).
The operating points for this experiment rely on three reasons:
($i$) The choice of $b_2$'s size ($d_2$) is a trade-off between filtering the noise to reduce its influence on blur estimation without loosing image details necessary to determine $b_1$. 
Hence, we pick the smallest defocus filters $d_2 \in \{7, 11\}\si{\px}$ that lead to stable blur estimation (cf. Fig.~\ref{fig:blurEstimationPhotonNoise+Defocus}).
($ii$) We consider small/medium motion blur $d_1 \in \{3, 7, 11\} \,\si{\px}$ so that the overall blur is still detectable by the CNN. 
($iii$) We focus on severe high/higher noise levels $\sigma \in \{10, 25\} \,\si{\DN}$.
We next evaluate $\widehat{\text{MTF}}(b_1) \approx \widehat{\text{MTF}}(b_1, b_2) ~/~ \text{MTF}^\text{GT}(b_2)$ with Table~\ref{tab:improvedBlurEstimationResults}.

We need to ensure three preconditions to divide $\text{MTF}^\text{GT}(b_2)$ from $\widehat{\text{MTF}}(b_1, b_2)$ for a meaningful result:
($i$) $\widehat{\text{MTF}}(b_1, b_2) \leq \text{MTF}^\text{GT}(b_2)$,
($ii$) $\text{MTF}^\text{GT}(b_2) > 0 + \epsilon$ and 
($iii$) $\widehat{\text{MTF}}(b_1, b_2) > 0 + \epsilon$, 
for all sampled frequencies.
We chose the control parameter $\epsilon = 0.05$ to avoid large quotients for small values, and omit frequencies that do not satisfy the conditions.
\begin{table}[t]
    \caption{\emph{Estimation of linear motion blur $b_1$} (Lin.~MB) on combined pipeline (Lin.~MB + DCSN + Defocus), using Udacity data. 
    The table reports mean absolute errors (MAE) of horizontal
    (H) and vertical (V) estimations, their average (AMAE), and their expected values ($\text{AMAE}^\text{Exp.}$ \eqref{eq:AMAE:expected}).}
    \label{tab:improvedBlurEstimationResults}
    \begin{adjustbox}{max width=1.0\linewidth}
    \setlength{\tabcolsep}{4pt}
    \begin{tabular}{cccccccc}
        \toprule
         \multicolumn{3}{c}{Corruption Levels} && \multicolumn{4}{c}{Error Metrics} \\ \cmidrule{1-3} \cmidrule{5-8}
         & $d_1 \,[\si{\px}]$ & $\sigma \,[\si{\DN}]$ & & MAE\,(H) & MAE\,(V) & AMAE & $\text{AMAE}^{\text{Exp.}}$\\
         \midrule
            \parbox[t]{0.1mm}{\multirow{6}{*}{\rotatebox[origin=c]{90}{$d_2 = \SI{7}{\px}$}}} & 3 & 10 && 2.9 & 2.0 & 2.5 & 16.2\\
            &  3 & 25 & &  3.5 & 2.1 & 2.8 & 16.2\\
            & 7 & 10 & &  12.3 & 7.7 & 10.0 & 10.8\\
            & 7 & 25 & &  11.6 & 8.8 & 10.2 & 10.8\\
            & 11 & 10 & &  11.1 & 8.4 & 9.7 & 9.8\\
            & 11 & 25 & &  14.1 & 10.1 & 12.1 & 9.8\\[1.2ex]
            \parbox[t]{0.1mm}{\multirow{6}{*}{\rotatebox[origin=c]{90}{$d_2 = \SI{11}{\px}$}}} & 3 & 10 & &  1.5 & 0.3 & 0.9 & 16.2\\
            & 3 & 25 & &  1.7 & 0.3 & 1.0 & 16.2\\
            & 7 & 10 & &  14.7 & 12.7 & 13.7 & 10.8\\
            & 7 & 25 & &  14.8 & 12.8 & 13.8 & 10.8\\
            & 11 & 10 & &  18.6 & 14.1 & 16.4 & 9.8\\
            & 11 & 25 & &  20.2 & 15.1 & 17.6 & 9.8\\
        \bottomrule
    \end{tabular}%
    \end{adjustbox}%
\end{table}

Table \ref{tab:improvedBlurEstimationResults} presents results in terms of MAE and AMAE scores \eqref{eq:amaedef}, and their expected values
\begin{equation}
\label{eq:AMAE:expected}
{\small
\text{AMAE}^\text{Exp.} \doteq \sqrt{\text{AMAE}(\widehat{\text{MTF}}(b_1))^2 + \text{AMAE}(\widehat{\text{MTF}}(b_2))^2}
}
\end{equation}
from the error propagation of $\widehat{\text{MTF}}(b_1)$ and $\widehat{\text{MTF}}(b_2)$.

We observe generally worse MAE scores in horizontal than in vertical image direction, which are in agreement with the already-mentioned slight motion blur in the moving direction on Udacity data (the moving direction is closest to the horizontal image axis; see Sec.~\ref{sec:experimentsBlurEstimation}).
It can be also seen that the higher the considered noise and blur levels, the worse the estimations of $b_1$.
The impact of higher noise, which relativizes with increasing $d_1$, is in agreement with the results of Fig.~\ref{fig:blurEstimationPhotonNoise+Defocus}.
Higher blur levels $d_1$ or $d_2$ increase the loss of information (where the MTF drops below zero) and thus worsen estimations of $b_1$.
This is also why the smaller defocus $d_2 = \SI{7}{\px}$ performs better (with results closer to their expected values) and smaller motion blurs $d_1$ are estimated more accurately (despite their higher expected values).
Moreover, the information loss causes the CNN to generally overestimate $d_1$, which in turn limits the estimation error for $d_1 = \SI{3}{\px}$ as its MTF values for the considered frequencies are already close to one.
All in all, a defocus filter with $d_2 = \SI{7}{\px}$ has been shown to be the best working solution to restore a blur estimation of $d_1$ in presence of high noise.

\emph{Summarizing}, additional defocus filtering suppresses noise so that estimation of preceding small or medium blur can be re-enabled for high sensor noise levels $\sigma \geq \SI{10}{\DN}$. 
This procedure is also suitable for a condition monitoring application as it can be applied in the background without changing the camera configuration.

\section{Blur direction dependence of YOLOv7}
\label{sec:yolov7}%
\begin{table}[t]
    \caption{\emph{YOLOv7 car detection performance as a function of blur direction, blur size, and noise level ($\sigma$)}.
    Rotation angles $\phi \in \{0,45,90,135 \} \, \si{\deg}$ are applied counter-clockwise.
    Red colours indicate row-wise outliers that violate monotonicity.
    }\label{tab:yolov7BlurDirectionPerformance}
    \centering
    \begin{adjustbox}{max width=1.0\linewidth}
    \begin{tabular}{llccccc}
        \toprule
        &&\multicolumn{5}{c}{Linear Motion Blur}\\
        \cmidrule(r){3-7} 
        \multicolumn{2}{l}{Size [$\si{px}$]} & 3 & 7 & 11 & 15 & 21%
        \\
        \multicolumn{2}{l}{Kernel} & \includegraphics[width=0.023\textwidth]{images/blurKernels/linearMotion/3.png}%
        & \includegraphics[width=0.023\textwidth]{images/blurKernels/linearMotion/7.png}%
        & \includegraphics[width=0.023\textwidth]{images/blurKernels/linearMotion/11.png}%
        & \includegraphics[width=0.023\textwidth]{images/blurKernels/linearMotion/15.png}%
        & \includegraphics[width=0.023\textwidth]{images/blurKernels/linearMotion/21.png}
        \\
        \midrule
        \parbox[t]{0.2mm}{\multirow{4}{*}{\rotatebox[origin=c]{90}{\tiny $\sigma = \SI{0}{\DN}$}}}%
            &0       & 50.31 & 49.68 & 43.52 & \textcolor{red}{44.38} & 27.17\\
            &45      & 50.83 & 49.97 & 48.19 & 48.03 & 38.81\\
            &90      & 50.80 & 50.56 & 49.02 & 41.57 & \textcolor{red}{43.57} \\
            &135     & 50.89 & 50.70 & \textcolor{red}{50.71}  & 41.60 & 37.49\\[1.2ex]
        \parbox[t]{0.2mm}{\multirow{4}{*}{\rotatebox[origin=c]{90}{\tiny $\sigma = \SI{5}{\DN}$}}}%
            &0       & 50.60 & 50.05 & 43.12 & \textcolor{red}{45.91} & 25.80\\
            &45      & 49.93 & 46.58 & \textcolor{red}{46.59} & 45.65 & 38.33\\
            &90      & 50.16 & 49.39 & 47.58 & 39.38 & \textcolor{red}{41.76}\\
            &135     & 49.60 & 48.48 & 48.42 & 39.00 & 37.76\\[1.2ex]
        \parbox[t]{0.2mm}{\multirow{4}{*}{\rotatebox[origin=c]{90}{\tiny $\sigma = \SI{10}{\DN}$}}}%
            &0       & 49.64 & 48.75 & 41.76 & \textcolor{red}{44.63} & 24.10\\
            &45      & 48.31 & 44.16 & \textcolor{red}{44.80} & 43.35 & 36.30\\
            &90      & 47.59 & 47.37 & 45.80 & 37.64 & \textcolor{red}{40.29}\\
            &135     & 48.13 & 46.45 & 45.95 & 37.84 & 36.26\\[1.2ex]
        \parbox[t]{0.2mm}{\multirow{4}{*}{\rotatebox[origin=c]{90}{\tiny $\sigma = \SI{15}{\DN}$}}}%
            &0       & 48.46 & 47.07 & 40.38 & \textcolor{red}{43.29} & 23.31\\
            &45      & 45.71 & 42.36 & \textcolor{red}{42.55} & 41.13 & 33.77\\
            &90      & 46.07 & 45.97 & 43.72 & 35.39 & \textcolor{red}{38.26}\\
            &135     & 45.37 & 45.14 & 42.94 & 35.52 & 32.65\\[1.2ex]
        \parbox[t]{0.2mm}{\multirow{4}{*}{\rotatebox[origin=c]{90}{\tiny $\sigma = \SI{20}{\DN}$}}}%
            &0       & 47.19 & 46.57 & 39.90 & \textcolor{red}{42.57} & 21.55\\
            &45      & 44.85 & 40.97 & 39.58 & 39.10 & 31.37\\
            &90      & 44.09 & 42.79 & 42.29 & 33.18 & \textcolor{red}{36.09}\\
            &135     & 44.13 & 43.18 & 41.29 & 33.56 & 30.53\\[1.2ex]
        \parbox[t]{0.2mm}{\multirow{4}{*}{\rotatebox[origin=c]{90}{\tiny $\sigma = \SI{25}{\DN}$}}}%
            &0       & 45.14 & 44.23 & 38.98 & \textcolor{red}{41.66} & 21.89\\
            &45      & 42.47 & 37.68 & 37.34 & 36.78 & 27.47\\
            &90      & 42.64 & 40.37 & 39.90 & 30.98 & \textcolor{red}{34.80}\\
            &135     & 42.43 & 41.35 & 38.58 & 30.59 & 28.49\\
        \bottomrule
    \end{tabular}%
    \end{adjustbox}
\end{table} 

The YOLOv4 car performance curves show non-linearities and non-monotonicity in terms of blur and noise (Fig.~\ref{fig:applicationMotionBlurPerformanceCurveResults}).
We examined the latest YOLOv7 object detector \cite{wang2023yolov7} for the same effect and found unexpected behavior when blur direction is incorporated as a third dimension.

From Tab.~\ref{tab:yolov7BlurDirectionPerformance}, it can be observed that car detection performance tends to increase for blur in horizontal image direction (compare values within a column for a fixed noise level).
The larger the kernel size, the more the performances vary across the different blur directions.
This influence of blur direction can even increase the detection performance as the blur size increases (see red values per row in Tab.~\ref{tab:yolov7BlurDirectionPerformance}).

We attribute this observation to the MS Coco dataset \cite{lin2014microsoft} used to train YOLOv7 \cite{wang2023yolov7}. 
Since natural motion blur occurs primarily for objects moving in horizontal image direction (where the angle between the camera and a scene object is most favourable for motion blur, cf.~\cite[p.~326]{bookJayaraman2009}), a dataset of natural images is likely to contain more examples of horizontal motion blur.
Thus, the training dataset may be biased, which would also affect YOLOv7.
Future studies could balance datasets for the directions in which motion blur occurs.

\section{Influence of Camera Parameters on Object Detection Performance}
\label{sec:objectDetectionPerformanceDependingOnCameraParameters}
\begin{figure}[t]
        \begingroup
            \pgfplotsset{every axis/.style={scale=0.55}}
            \pgfplotsset{cycle list/Set1, cycle multiindex* list={
                    mark list*\nextlist
                    Set1\nextlist
                }
            }
            \begin{tikzpicture}
 \scriptsize
    \begin{axis}[
        name=axis1,
        xshift=+10mm,
        tick pos=left,
        title style={text width=18em, text centered},
        title={\textbf{Lin. MB + Sensor Noise: \\ YOLOv4 Car}},
        x grid style={white!69.0196078431373!black},
        xlabel={ISO Gain $A_\text{ISO}$ [$\times 10^2$]},
        xmin=-0.5, xmax=5.5,
        xtick style={color=black},
        xtick={0,1,2,3,4,5},
        xticklabels={1, 5, 10, 15, 20, 25},
        y grid style={white!69.0196078431373!black},
        ylabel={Exp. Time $t_\text{exp}$ [\si{\milli\second}]},
        ymin=-0.5, ymax=6.5,
        ytick style={color=black},
        ytick={0.0,1.5,2.5,3.5,4.5,5.5,6.5},
        yticklabels={1, 3, 7, 11, 13, 15, 21}
        ]
        \addplot graphics [includegraphics cmd=\pgfimage,xmin=-1.5, xmax=6.3, ymin=-1.5, ymax=7.6] {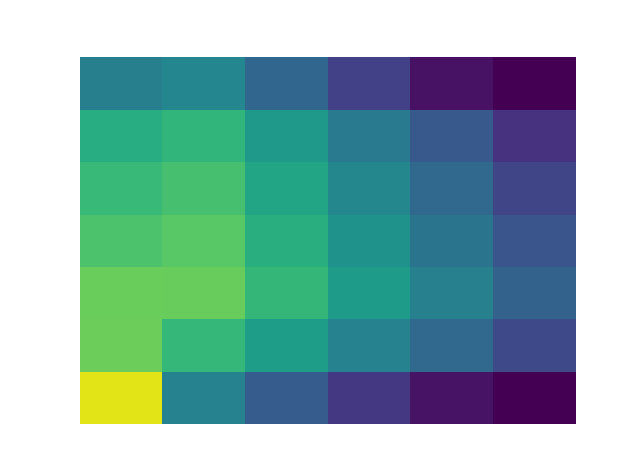};
        \draw (axis cs:0,6) node[
          text=white,
          rotate=0.0
        ]{19.5};
        \draw (axis cs:1,6) node[
          text=white,
          rotate=0.0
        ]{19.9};
        \draw (axis cs:2,6) node[
          text=white,
          rotate=0.0
        ]{18.2};
        \draw (axis cs:3,6) node[
          text=white,
          rotate=0.0
        ]{16.4};
        \draw (axis cs:4,6) node[
          text=white,
          rotate=0.0
        ]{14.5};
        \draw (axis cs:5,6) node[
          text=white,
          rotate=0.0
        ]{12.6};
        \draw (axis cs:0,5) node[
          text=white,
          rotate=0.0
        ]{22.0};
        \draw (axis cs:1,5) node[
          text=white,
          rotate=0.0
        ]{22.4};
        \draw (axis cs:2,5) node[
          text=white,
          rotate=0.0
        ]{20.9};
        \draw (axis cs:3,5) node[
          text=white,
          rotate=0.0
        ]{19.2};
        \draw (axis cs:4,5) node[
          text=white,
          rotate=0.0
        ]{17.5};
        \draw (axis cs:5,5) node[
          text=white,
          rotate=0.0
        ]{15.7};
        \draw (axis cs:0,4) node[
          text=white,
          rotate=0.0
        ]{22.7};
        \draw (axis cs:1,4) node[
          text=white,
          rotate=0.0
        ]{23.1};
        \draw (axis cs:2,4) node[
          text=white,
          rotate=0.0
        ]{21.6};
        \draw (axis cs:3,4) node[
          text=white,
          rotate=0.0
        ]{19.9};
        \draw (axis cs:4,4) node[
          text=white,
          rotate=0.0
        ]{18.3};
        \draw (axis cs:5,4) node[
          text=white,
          rotate=0.0
        ]{16.6};
        \draw (axis cs:0,3) node[
          text=white,
          rotate=0.0
        ]{23.2};
        \draw (axis cs:1,3) node[
          text=white,
          rotate=0.0
        ]{23.5};
        \draw (axis cs:2,3) node[
          text=white,
          rotate=0.0
        ]{22.1};
        \draw (axis cs:3,3) node[
          text=white,
          rotate=0.0
        ]{20.5};
        \draw (axis cs:4,3) node[
          text=white,
          rotate=0.0
        ]{18.9};
        \draw (axis cs:5,3) node[
          text=white,
          rotate=0.0
        ]{17.3};
        \draw (axis cs:0,2) node[
          text=white,
          rotate=0.0
        ]{23.9};
        \draw (axis cs:1,2) node[
          text=white,
          rotate=0.0
        ]{23.9};
        \draw (axis cs:2,2) node[
          text=white,
          rotate=0.0
        ]{22.5};
        \draw (axis cs:3,2) node[
          text=white,
          rotate=0.0
        ]{21.0};
        \draw (axis cs:4,2) node[
          text=white,
          rotate=0.0
        ]{19.5};
        \draw (axis cs:5,2) node[
          text=white,
          rotate=0.0
        ]{18.0};
        \draw (axis cs:0,1) node[
          text=white,
          rotate=0.0
        ]{24.0};
        \draw (axis cs:1,1) node[
          text=white,
          rotate=0.0
        ]{22.6};
        \draw (axis cs:2,1) node[
          text=white,
          rotate=0.0
        ]{21.1};
        \draw (axis cs:3,1) node[
          text=white,
          rotate=0.0
        ]{19.7};
        \draw (axis cs:4,1) node[
          text=white,
          rotate=0.0
        ]{18.2};
        \draw (axis cs:5,1) node[
          text=white,
          rotate=0.0
        ]{16.8};
        \draw (axis cs:0,0) node[
          text=white,
          rotate=0.0
        ]{26.3};
        \draw (axis cs:1,0) node[
          text=white,
          rotate=0.0
        ]{19.7};
        \draw (axis cs:2,0) node[
          text=white,
          rotate=0.0
        ]{17.6};
        \draw (axis cs:3,0) node[
          text=white,
          rotate=0.0
        ]{16.0};
        \draw (axis cs:4,0) node[
          text=white,
          rotate=0.0
        ]{14.5};
        \draw (axis cs:5,0) node[
          text=white,
          rotate=0.0
        ]{13.1};
\end{axis}

\begin{axis}[
        name=axis2,
        at=(axis1.right of south east),
        xshift=+0mm,
        tick pos=left,
        title style={text width=18em, text centered},
        title={\textbf{Lin. MB + Sensor Noise: \\ Faster R-CNN Pedestrian}},
        x grid style={white!69.0196078431373!black},
        xlabel={ISO Gain $A_\text{ISO}$ [$\times 10^2$]},
        xmin=-0.5, xmax=5.5,
        xtick style={color=black},
        xtick={0,1,2,3,4,5},
        xticklabels={1, 5, 10, 15, 20, 25},
        y grid style={white!69.0196078431373!black},
        ymin=-0.5, ymax=6.5,
        ytick style={color=black},
        ytick={0.0,1.5,2.5,3.5,4.5,5.5,6.5},
        yticklabels=\empty
        ]
        \addplot graphics [includegraphics cmd=\pgfimage,xmin=-1.5, xmax=6.3, ymin=-1.5, ymax=7.6] {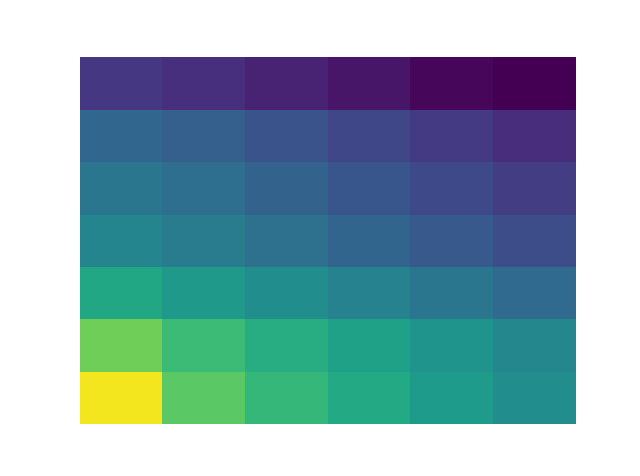};
        \draw (axis cs:0,6) node[
          text=white,
          rotate=0.0
        ]{9.7};
        \draw (axis cs:1,6) node[
          text=white,
          rotate=0.0
        ]{9.1};
        \draw (axis cs:2,6) node[
          text=white,
          rotate=0.0
        ]{8.0};
        \draw (axis cs:3,6) node[
          text=white,
          rotate=0.0
        ]{6.9};
        \draw (axis cs:4,6) node[
          text=white,
          rotate=0.0
        ]{5.7};
        \draw (axis cs:5,6) node[
          text=white,
          rotate=0.0
        ]{4.6};
        \draw (axis cs:0,5) node[
          text=white,
          rotate=0.0
        ]{14.7};
        \draw (axis cs:1,5) node[
          text=white,
          rotate=0.0
        ]{13.8};
        \draw (axis cs:2,5) node[
          text=white,
          rotate=0.0
        ]{12.6};
        \draw (axis cs:3,5) node[
          text=white,
          rotate=0.0
        ]{11.3};
        \draw (axis cs:4,5) node[
          text=white,
          rotate=0.0
        ]{10.1};
        \draw (axis cs:5,5) node[
          text=white,
          rotate=0.0
        ]{8.8};
        \draw (axis cs:0,4) node[
          text=white,
          rotate=0.0
        ]{16.4};
        \draw (axis cs:1,4) node[
          text=white,
          rotate=0.0
        ]{15.4};
        \draw (axis cs:2,4) node[
          text=white,
          rotate=0.0
        ]{14.2};
        \draw (axis cs:3,4) node[
          text=white,
          rotate=0.0
        ]{12.9};
        \draw (axis cs:4,4) node[
          text=white,
          rotate=0.0
        ]{11.6};
        \draw (axis cs:5,4) node[
          text=white,
          rotate=0.0
        ]{10.3};
        \draw (axis cs:0,3) node[
          text=white,
          rotate=0.0
        ]{18.2};
        \draw (axis cs:1,3) node[
          text=white,
          rotate=0.0
        ]{17.1};
        \draw (axis cs:2,3) node[
          text=white,
          rotate=0.0
        ]{15.8};
        \draw (axis cs:3,3) node[
          text=white,
          rotate=0.0
        ]{14.5};
        \draw (axis cs:4,3) node[
          text=white,
          rotate=0.0
        ]{13.1};
        \draw (axis cs:5,3) node[
          text=white,
          rotate=0.0
        ]{11.8};
        \draw (axis cs:0,2) node[
          text=white,
          rotate=0.0
        ]{22.2};
        \draw (axis cs:1,2) node[
          text=white,
          rotate=0.0
        ]{20.8};
        \draw (axis cs:2,2) node[
          text=white,
          rotate=0.0
        ]{19.3};
        \draw (axis cs:3,2) node[
          text=white,
          rotate=0.0
        ]{17.9};
        \draw (axis cs:4,2) node[
          text=white,
          rotate=0.0
        ]{16.4};
        \draw (axis cs:5,2) node[
          text=white,
          rotate=0.0
        ]{15.0};
        \draw (axis cs:0,1) node[
          text=white,
          rotate=0.0
        ]{27.5};
        \draw (axis cs:1,1) node[
          text=white,
          rotate=0.0
        ]{24.8};
        \draw (axis cs:2,1) node[
          text=white,
          rotate=0.0
        ]{23.1};
        \draw (axis cs:3,1) node[
          text=white,
          rotate=0.0
        ]{21.5};
        \draw (axis cs:4,1) node[
          text=white,
          rotate=0.0
        ]{20.0};
        \draw (axis cs:5,1) node[
          text=white,
          rotate=0.0
        ]{18.5};
        \draw (axis cs:0,0) node[
          text=white,
          rotate=0.0
        ]{33.4};
        \draw (axis cs:1,0) node[
          text=white,
          rotate=0.0
        ]{26.6};
        \draw (axis cs:2,0) node[
          text=white,
          rotate=0.0
        ]{24.3};
        \draw (axis cs:3,0) node[
          text=white,
          rotate=0.0
        ]{22.5};
        \draw (axis cs:4,0) node[
          text=white,
          rotate=0.0
        ]{20.9};
        \draw (axis cs:5,0) node[
          text=white,
          rotate=0.0
        ]{19.3};
\end{axis}
        
  \end{tikzpicture}
        \endgroup
    \vspace{-2ex}
    \caption{\emph{Influence of exposure time ($t_\text{exp}$) and ISO gain ($A_\text{ISO}$) on object detection performance (AP).}
    Regressed from the curves in Fig.~\ref{fig:applicationMotionBlurPerformanceCurveResults} using \eqref{eq:detectionPerformanceByExposureTimeAndISOGain} with coefficients from Tab.~\ref{tab:performanceProfileFittingParameters}.
    }
    \label{fig:performanceProfileDataFit}
\end{figure}
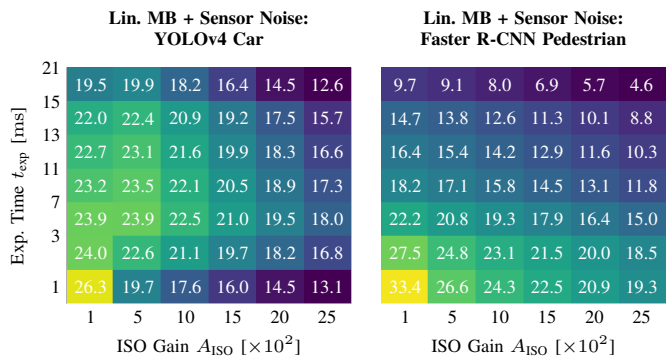
\begin{table}[t]
    \caption{\emph{Coefficients to fit \eqref{eq:detectionPerformanceByExposureTimeAndISOGain} to the curves in Fig.~\ref{fig:applicationMotionBlurPerformanceCurveResults}.} $\chi^2$ ($\downarrow$) denotes the goodness-of-fit and $R^2$ ($\uparrow$) the coefficient of determination score.}
    \label{tab:performanceProfileFittingParameters}
    \begin{adjustbox}{max width=1.0\linewidth}
    \begin{tabular}{ccccc}
        \toprule
        & \multicolumn{2}{c}{YOLOv4 Car} & \multicolumn{2}{c}{Faster R-CNN Pedestrian} \\ 
        \cmidrule{2-3}   \cmidrule{4-5}
         Param. & Value & Std. Dev. & Value & Std. Dev.\\
         \midrule
            $c_1$ &        \num{-5.5862e-05}  &  \num{2.4493e-05} & \num{4.0763e-05} & \num{2.4130e-05} \\
            $c_2$ & \num{-2.6957e-03} &  \num{3.1564e-4} &  \num{-3.1039e-03} &  \num{3.1096e-04}\\
            $c_3$ &  \num{-1.1775e+00}  &  \num{1.915e-01} &  \num{-4.9352e-01} & \num{1.8868e-01}\\
            $c_4$ &        \num{6.6093e+00}  &  \num{1.3270e+00} &  \num{-2.7007e+00} & \num{1.3074e+00}\\
            $c_5$ &        \num{1.6871e+01}  &  \num{2.4239e+00} &  \num{3.3140e+01} & \num{2.3879e+00}\\
            $c_6$ &        \num{9.8229e+02}  &  \num{1.2080e+02} &  \num{7.7306e+02} & \num{1.1901e+02}\\
            $c_7$ &        \num{-2.9259e+02}  &  \num{5.6069e+01} &  \num{-6.4604e+01} & \num{5.5238e+01}\\
            $c_8$ &        \num{-2.6307} &   \num{1.3947} &  \num{-3.2796} & \num{1.3740}\\
            \midrule
            $\chi^2$ & 17.4500 & - & 16.9363 & -\\
            $R^2$ & 0.9602 & - & 0.9898 & -\\
        \bottomrule
    \end{tabular}%
    \end{adjustbox}%
\end{table}

Object detection AP performances can be expressed as a function of exposure time ($t_\text{exp}$) and ISO gain ($A_\text{ISO}$) camera parameters, if static camera and environmental conditions can be assumed (i.e., constant blur and noise statistics). 
To this end, we apply a polynomial least-squares regression to the IOPCs from Fig.~\ref{fig:applicationMotionBlurPerformanceCurveResults} with the constraint to keep the degree of the polynomial low (to avoid overfitting) while accounting for the IOPSs' non-linearity and non-monotonicity.
We identify the different components of the polynomial manually by mapping the slopes of the IOPCs into respective terms. 
This yields \eqref{eq:detectionPerformanceByExposureTimeAndISOGain}:
\begin{equation}
\label{eq:detectionPerformanceByExposureTimeAndISOGain2}
    \widehat{\text{AP}} = 
    c_1 xy + c_2 x + c_3 y + c_4 \sqrt{y} + c_5
    + \frac{c_6}{x y} + \frac{c_7}{x} + \frac{c_8}{y}  
\end{equation}
with $x = \lambda_1 A_\text{ISO}$, $y = \lambda_2 t_\text{exp}$, and regression coefficients $c_1, \dots, c_8$ (values in Tab.~\ref{tab:performanceProfileFittingParameters}).
Note that $\lambda_1$ and $\lambda_2$ must be calibrated beforehand to match occurring blur and noise statistics to the camera parameters.
We calibrated both so that $t_\text{exp} = \SI{1}{\milli\second}$ and $A_\text{ISO} = 100$ leads to $\widehat{\text{MTF}} = 1.0$ and $\hat\sigma = \SI{1}{\DN}$.

The corresponding IOPCs as a function of the camera parameters are provided in Fig.~\ref{fig:performanceProfileDataFit}.
We can see that these IOPCs match well with those in Fig.~\ref{fig:applicationMotionBlurPerformanceCurveResults}.
This is also reflected by the goodness-of-fit ($\chi^2$) \cite[pp.~595--596]{poissonToGaussDistr} and coefficient of determination ($R^2$) \cite[pp.~484--485]{poissonToGaussDistr} statistics in Tab.~\ref{tab:performanceProfileFittingParameters}.
The omission of any polynomial term leads to significant degradations of these scores.
Still, the non-monotonic part of the YOLOv4 curve could be improved, but at the expense of higher degree factors in \eqref{eq:detectionPerformanceByExposureTimeAndISOGain2}.

\ifarxiv

\fi

\vspace{-0.75cm}
\begin{IEEEbiography}[{\includegraphics[width=1in,height=1.25in,clip,keepaspectratio]{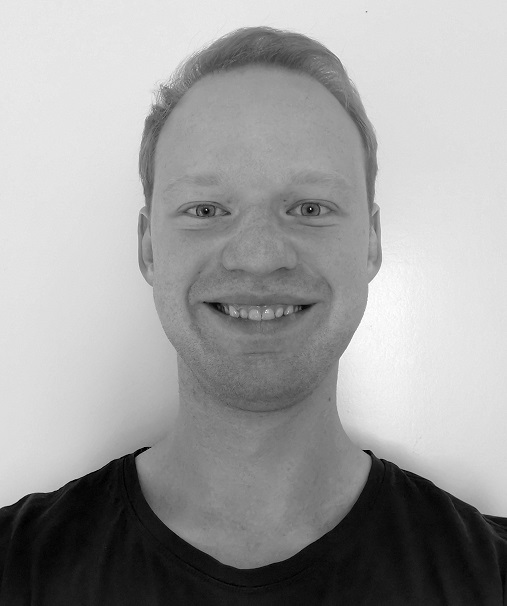}}]{Maik Wischow} is a PhD student at the German Aerospace Center (DLR) and the Technische Universit\"at Berlin, where he also received the M.Sc. degree in Computer Science in 2019.
His research interests include computer vision, deep neural networks, optical sensor systems, optics, stereo image processing and sensor fusion.
\end{IEEEbiography}
\begin{IEEEbiography}[{\includegraphics[width=1in,height=1.25in,clip,keepaspectratio]{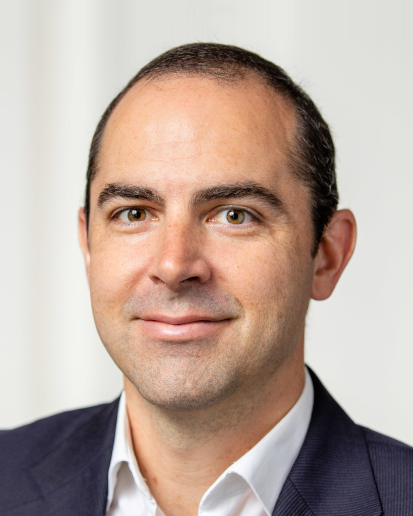}}]{Guillermo Gallego} (SM'19) is Associate Professor at the Technische Universit\"at Berlin, in the Dept. of Electrical Engineering and Computer Science, and at the Einstein Center Digital Future, both in Berlin, Germany.
He received the PhD degree in Electrical and Computer Engineering from the Georgia Institute of Technology, USA, in 2011, supported by a Fulbright Scholarship.
From 2011 to 2014 he was a Marie Curie researcher with Universidad Politecnica de Madrid, Spain, and from 2014 to 2019 he was a postdoctoral researcher at the Robotics and Perception Group, 
University of Zurich, Switzerland.
His research interests include robotics, computer vision, signal processing, optimization and geometry. 
\end{IEEEbiography}
\begin{IEEEbiography}[{\includegraphics[width=1in,height=1.25in,clip,keepaspectratio]{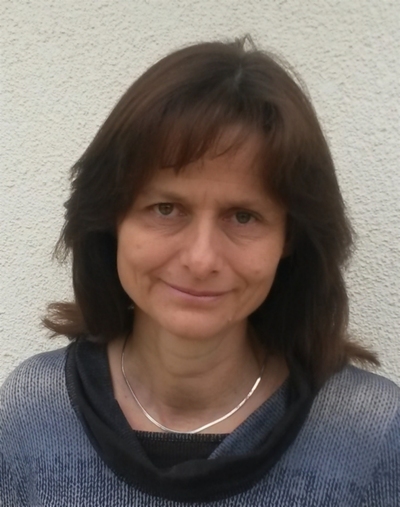}}]{Ines Ernst} received the diploma degree in mathematics from the Technische Universität Dresden.
Until 2001 she worked at the Institute for Computer Architecture and Software Technology of the German National Research Center for Information Technology (GMD FIRST).
She joined the German Aerospace Center (DLR) in 2002 and is currently a research associate at the Institute of Optical Sensor Systems.
Her research interests include computer vision, stereo image processing and 3D reconstruction, deep neural networks, hardware acceleration.
\end{IEEEbiography}
\begin{IEEEbiography}[{\includegraphics[width=1in,height=1.5in,clip,keepaspectratio]{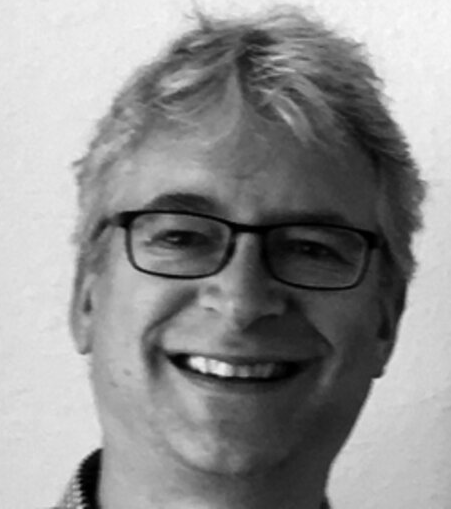}}]{Anko B\"orner} received the degree in electrical engineering from the University of Ilmenau, and the PhD degree from the German Aerospace Center (DLR).
He was a Postdoctoral Researcher with the University of Zurich. 
He is currently the Head of the Real-Time Data Processing Department, DLR. 
He is author of various SCI, EI, and Scopus indexed journals and international conferences. 
His research interests include stereo image processing, deep neural networks, simultaneous localization and modeling (SLAM), and 3D reconstruction.
\end{IEEEbiography}

\vfill

\end{document}